\documentclass[11pt]{article}

\usepackage{arxiv}
\usepackage{amsmath,amssymb,amsthm,booktabs}
\usepackage[numbers]{natbib}
\usepackage{hyperref}
\usepackage{xcolor}
\usepackage{titlesec}
\usepackage{tcolorbox}

\titleformat{\section}{\normalfont\large\bfseries}{\thesection}{0.6em}{}
\titleformat{\subsection}{\normalfont\bfseries}{\thesubsection}{0.6em}{}

\definecolor{novemberrevision}{RGB}{0,105,95}

\newcommand{\nrev}[1]{{\color{novemberrevision}#1}}
\definecolor{aaaistrengthening}{RGB}{135,35,70}

\newcommand{\slin}{s^{\mathrm{lin}}_\caus}
 
\newtheorem{theorem}{Theorem}
\newtheorem{proposition}{Proposition}
\newtheorem{corollary}{Corollary}
\newtheorem{lemma}{Lemma}
\newtheorem{assumption}{Assumption}
\theoremstyle{definition}
\newtheorem{definition}{Definition}
\newtheorem{remark}{Remark}

\newcommand{\E}{\mathbb{E}}
\newcommand{\R}{\mathbb{R}}
\newcommand{\KL}{\mathrm{KL}}

\newcommand{\supp}{\mathrm{supp}}
\newcommand{\act}{\mathrm{act}}
\newcommand{\dec}{\mathrm{dec}}
\newcommand{\caus}{\mathrm{caus}}
\newcommand{\iden}{\iota}
\newcommand{\1}{\mathbf 1}
\newcommand{\uu}{\mathbf u}
\newcommand{\vv}{\mathbf v}

\DeclareMathOperator*{\argmin}{arg\,min}

\newcommand{\Dis}{\mathrm{D}}                        
\newcommand{\opn}[1]{\lVert #1\rVert_{\mathrm{op}}}  
\newcommand{\Mhat}{\widehat{M}}                      
\newcommand{\emin}{e_{\min}}                         
\newcommand{\Ncell}{N_{\mathrm{cell}}}
\newcommand{\etafr}{\eta_{\mathrm{fr}}}              
\newcommand{\Cov}{\mathrm{Cov}}
\newcommand{\Var}{\mathrm{Var}}
\newcommand{\corr}{\mathrm{corr}}

\theoremstyle{definition}

\providecommand{\Dis}{\mathrm{D}}                        
\providecommand{\opn}[1]{\lVert #1\rVert_{\mathrm{op}}}  
\providecommand{\Mhat}{\widehat{M}}                      
\providecommand{\emin}{e_{\min}}                         
\providecommand{\Ncell}{N_{\mathrm{cell}}}
\providecommand{\etafr}{\eta_{\mathrm{fr}}}              
\providecommand{\Cov}{\mathrm{Cov}}
\providecommand{\Var}{\mathrm{Var}}
\providecommand{\corr}{\mathrm{corr}}

\title{When Are Two SAE Latents the Same Feature?}
\author{Alexandra Suvorikova\\
Independent researcher
\And Ilgam Latypov\\
MSU Institute for Artificial Intelligence\\
\And Anton Trukhan\\
Independent researcher
\And Yuriy Dorn \\
MSU Institute for Artificial Intelligence\\
}

\date{}

\begin{document}
\maketitle

\begin{abstract}
Comparing sparse autoencoders (SAEs) requires connecting feature correspondence, differences between whole dictionaries, and uncertainty from finite evaluation data. We develop a framework that connects these tasks through an energy-weighted optimal-transport plan under a specified similarity geometry. The same plan defines a dictionary-level discrepancy, decomposes it into latent contributions, and quantifies how each latent's mass is distributed across counterparts. We characterize relationships between decoder, activation, and first-order causal similarities; idealized models clarify the interpretation of splitting and absorption. For fixed trained SAEs, we derive joint asymptotic inference for transport costs and fractions, accounting for activation similarities and both energy marginals estimated from the same documents. A joint document bootstrap gives asymptotically valid simultaneous confidence bands. Experiments across four language models locate cross-seed discrepancy and separate context-dependent changes.
\end{abstract}

\section{Introduction}\label{sec:intro}

Sparse autoencoders (SAEs) decompose neural activations into sparse
learned components, enabling comparisons across training runs, widths,
and model checkpoints
\citep{bricken2023monosemanticity,lieberum2024gemmascope,leask2025canonical}.
We use latent to denote a learned encoder-decoder component, and feature to refer to a property of the model’s representations or behavior that such components may capture. Deciding whether two latents represent the same feature requires specifying what is compared: decoder directions, activation patterns,
effects on model outputs, or natural-language explanations
\citep{song2025consistency,paulo2025different,marks2024circuits,paulo2024automatically}.
These notions can disagree. Moreover, a feature may be split across
several latents or absorbed into a more specific one
\citep{bussmann2025matryoshka,chanin2024absorption}, so correspondence must
accommodate different allocations of information across dictionaries.

A useful comparison should quantify differences between entire
dictionaries, locate the latent correspondences accounting for those
differences, and distinguish them from sampling variability. An aggregate
score alone does not locate disagreement; separate pairwise similarities
do not account for how all latents share the available counterparts.
The evaluated inputs also matter: the same trained dictionaries may
allocate their activation energy differently across contexts.

We connect these tasks through a single energy-weighted transport plan.
Its cost measures dictionary-level discrepancy, its cost decomposition
locates the contributing latents, and its transported fractions quantify
how each latent's mass is distributed across counterparts. This gives
a quantitative answer to feature correspondence under an explicit
similarity geometry, input distribution, and regularization level.

\paragraph{Our contributions.}
\begin{itemize}
\item \textbf{A framework connecting dictionary and latent comparison.}
We couple normalized activation energies under a specified similarity
geometry. One plan gives a whole-dictionary discrepancy, its exact
decomposition into latent contributions, and allocation fractions for
individual latents or specified groups (Section~\ref{sec:stab}).
\item \textbf{Geometric and structural interpretation.}
We express decoder, activation, and causal similarities as Hilbert-space
cosines and characterize their relationships, including factorization
and bounds for first-order causal similarity (Section~\ref{sec:setup}).
Idealized models establish splitting shares and the limiting
absorbed-energy interpretation of transport fractions.
\item \textbf{Joint inference from evaluation documents.}
For activation or fixed decoder costs with activation-energy weights,
we derive explicit influence functions, consistent covariance
estimators, and asymptotic intervals for costs and fractions, including
dependence between SAE pairs (Section~\ref{sec:fss}). A joint document
bootstrap gives simultaneous bands and full-plan error radii
(Appendix~\ref{ssec:bootstrap}).
\item \textbf{Sources of disagreement across trained SAEs.}
Across four language models, we compare independently initialized SAEs
trained on identical data within each setting. An ablation with fixed
activation costs isolates the role of energy weighting; a context contrast is decomposed
into changes in weights and co-activation; and selected latent
contributions are evaluated on held-out documents. Groups selected by
tuning energy also reveal substantial evaluation cost among
low-energy latents across larger dictionaries (Section~\ref{sec:exp}).
\end{itemize}

\subsection{Related Work}\label{sec:related}

\paragraph{Sparse autoencoders and their pathologies.}
SAEs recover interpretable directions from language-model activations at scale
\citep{bricken2023monosemanticity,cunningham2024sae,templeton2024scaling,gao2024scaling,rajamanoharan2024jumprelu,lieberum2024gemmascope},
with dictionary structure varying across training configurations: features split across widths and seeds
\citep{leask2025canonical,bussmann2025matryoshka}, general features are
absorbed into specific ones \citep{chanin2024absorption}, correlated features
distort narrow dictionaries \citep{chanin2025hedging}, and features can have multidimensional structure \citep{engels2025multidim}. Recent theory explains
several of these phenomena as properties of the dictionary-learning objective \citep{dorrell2026optimality,cui2025limits,tang2025sdl},
and the representational role of dense latents has been studied
\citep{sun2025dense}. \emph{Our splitting and absorption models idealize
these phenomena}, and our comparison of sparse reconstruction objectives
builds on the optimality viewpoint of \citet{dorrell2026optimality}.

\paragraph{Comparing features across runs, widths, and layers.}
\citet{leask2025canonical} match latents across widths by decoder cosine and
find no canonical units; \citet{paulo2025different} document that identically
trained SAEs learn different features; \citet{song2025consistency} argue that
feature consistency should be a primary evaluation target, and
\citet{jin2026c2r} train explicitly for it. Cross-layer analogues match decoder
parameters by nearest-cosine or linear assignment
\citep{balagansky2024saematch,laptev2025featureflow}, crosscoders sidestep
matching by tying the dictionaries \citep{lindsey2024crosscoders}, and
universality studies compare representation spaces across models
\citep{lan2024universal,huh2024platonic}. \emph{These approaches motivate complementary comparisons at the
levels of individual latents, groups of latents, and whole dictionaries.}

\paragraph{Optimal transport.}
Optimal transport is a standard tool for aligning models and representations:
model fusion and rebasing
\citep{singh2020fusion,ainsworth2022gitrebasin,imfeld2024transformer},
embedding alignment \citep{alvarezmelis2018gromov,memoli2011gromov}, and the
computational machinery we use
\citep{cuturi2013sinkhorn,peyre2019computational,villani2009optimal}. \citet{cao2026semantic} use OT between activation-weighted distributions
of hidden states to define a distance for each feature pair; their feature-matching rule selects a nearest feature under that distance. \emph{Our transport matrix is indexed by pairs of latents. Each entry records the normalized activation energy allocated from one latent to another, subject to both dictionaries' marginal constraints.} OT soft matching \citep{khosla2024soft} couples two populations of units
with uniform marginals, and \citet{longon2026superposition} apply it to SAE latents of differently seeded networks. Uniform-mass OT is a special case of our comparison.

\section{Similarity Measures and Their Relationships}\label{sec:setup}

Let $x\sim\mathcal D$ be an input to a fixed pretrained model, and let
$a(x)\in\R^n$ denote its activation at the location used to train the SAEs,
such as a residual stream, an MLP output, or an attention-head output.

The SAE encoder maps $a(x)$ to nonnegative activations
$h(x)=(h_1(x),\dots,h_m(x))\in\R^m_{\ge0}$. Its decoder reconstructs
$\hat a(x)=Dh(x)+b=\sum_j h_j(x)d_j+b$, where $d_j\in\R^n$ is column $j$
of $D$. We call the pair $\ell_j=(h_j,d_j)$ a \emph{latent}:
$h_j(x)$ specifies its activation on input $x$, and $h_j(x)d_j$ is its contribution to the reconstruction.

We compare two SAEs $\mathcal A,\mathcal B$ trained at the same model
location, so their decoder columns lie in the same space $\R^n$.
To compare their effects on the model output, let $F:\R^n\to\R^V$ map
this activation to an output of interest, such as logits. Write
$J(x)=\nabla F(a(x))$ for its Jacobian. If $F(a)=Wa+b$ is affine, then
$J\equiv W$ is constant.

\begin{definition}[Similarity kernels]\label{def:kernels}
Decoder columns have unit norm; we rescale the corresponding
activations inversely when normalizing a column, preserving the
reconstruction.
For $j\in \mathcal A,\ k\in \mathcal B$:
\begin{itemize}
    \item \emph{decoder} $s_\dec(j,k):=\langle d_j,d_k\rangle$ \citep{bricken2023monosemanticity,leask2025canonical,paulo2025different};
    \item \emph{activation} (uncentered co-activation) $\rho_\act(j,k):=\frac{\E[h_jh_k]}{\sqrt{\E[h_j^2]\E[h_k^2]}}\in[0,1]$ \citep{song2025consistency};
    \item \emph{causal} $\hat s_\caus(j,k):=\E\frac{\langle e_j,e_k\rangle}{\sqrt{\E\|e_j\|^2\E\|e_k\|^2}}$ with $e_j(x):=F(a(x)-h_j(x)d_j)-F(a(x))$ \citep{marks2024circuits,makelov2024principled};
    \item \emph{linearized causal} $\slin(j,k)$, the same normalized
    inner product with $e_j$ replaced by its first-order part
    $q_j(x):=-h_j(x)J(x)d_j$ (Assumption~\ref{ass:firstorder}).
\end{itemize}
\end{definition}
Agreement between natural-language explanations is another notion
of similarity \citep{bills2023neurons,paulo2024automatically}, but it is out of the scope of the current study. The three kernels below compare decoder vectors, activation
functions, and output-change functions, respectively, as cosines in
different Hilbert spaces.
\begin{lemma}[Three kernels are Hilbert-space cosines]\label{lem:chordal}
The decoder, activation, and causal kernels of Definition~\ref{def:kernels} are
inner products of unit embeddings:
\begin{align*}
\Phi_\dec(j)&=d_j\in\R^n,\\
\Phi_\act(j)&=h_j/\|h_j\|_{L^2(\mathcal D)}\in L^2(\mathcal D),\\
\Phi_\caus(j)&=e_j/\|e_j\|_{L^2(\mathcal D;\R^V)}\in L^2(\mathcal D;\R^V).
\end{align*}
Consequently each of the three satisfies
$\|\Phi(j)-\Phi(k)\|^2=2-2s(j,k)\in[0,4]$.
\end{lemma}
Appendix~\ref{app:proofs} contains the proof.
The decoder and activation similarity can vary independently even
under exact reconstruction. 
\begin{proposition}[Decoder and activation similarity are mutually unconstrained]\label{thm:t1uncon}
There are exact-reconstruction SAE representations with $s_\dec(j,k)=1$ and
$\rho_\act(j,k)=0$, and exact-reconstruction SAE representations with
$s_\dec(j,k)=0$ and $\rho_\act(j,k)=1$. 
\end{proposition}
Appendix~\ref{app:proofs} contains the proof.
To relate causal similarity to these two quantities, we approximate
the output change caused by removing a latent contribution using the
Jacobian of $F$ \citep{marks2024circuits,makelov2024principled}.

\begin{assumption}[First-order approximation of the output change]\label{ass:firstorder}
The map $F$ is differentiable and the ablation steps
$h_j(x)d_j$ are small
enough that $e_j(x)=-h_j(x)J(x)d_j+O(h_j^2)$ with Jacobian
$J(x):=\nabla F(a(x))$. 
\end{assumption}
Proposition~\ref{thm:causal} concerns the linearized kernel $\slin$.
Under the additional smoothness and nondegeneracy assumptions of
Proposition~\ref{prop:causal-path} (Appendix~\ref{app:envelope}), an explicit
remainder bound controls its difference from the exact kernel
$\hat s_\caus$. The two coincide when $F$ is globally affine.
\begin{proposition}[Factorization and bound for linearized causal similarity]\label{thm:causal}
Under Assumption~\ref{ass:firstorder}:
\begin{itemize}
\item \emph{(Factorization.)} If $J(x)\equiv J$ is constant,
then
\[
\slin(j,k)=\rho_\act(j,k)\cdot\cos_M(d_j,d_k),
\]
where
\[
\cos_M(d_j,d_k):=\frac{\langle Jd_j,Jd_k\rangle}{\|Jd_j\|\|Jd_k\|}
\]
is the cosine between the transformed decoder vectors $Jd_j,Jd_k$,
induced by $M:=J^\top J$. 
\item \emph{(Co-activation bound.)} Since $|\cos_M|\le1$,
$|\slin(j,k)|\le\rho_\act(j,k)$ for constant $J$. For an
input-dependent Jacobian $J(x)$, define the Jacobian-weighted co-activation
$\rho_J(j,k):=N_{jk}\big/\sqrt{N_{jj}N_{kk}}\in[0,1]$, where
\[
N_{jk}:=\E\big[\|J(x)d_j\|\,\|J(x)d_k\|\,h_j(x)h_k(x)\big];
\]
then $|\slin(j,k)|\le\rho_J(j,k)$, and $\rho_J$ collapses to $\rho_\act$
when $J$ is constant. 
\item \emph{(Orthogonal decoders.)}
For constant $J$, orthogonal decoders can satisfy
$Jd_j=Jd_k\ne0$, giving $s_\dec=0$ and $\slin=\rho_\act$.
\item \emph{(Affine case.)} For affine $F$, $\hat s_\caus=\slin$.
\end{itemize}
\end{proposition}
Appendix~\ref{app:proofs} provides the proof.
Decoder similarity compares directions in activation space,
activation similarity compares activation patterns across inputs,
and causal similarity compares changes in model outputs.
The map $F$ can change angles between decoder directions.
For nonlinear $F$, Proposition~\ref{prop:causal-path}
(Appendix~\ref{app:envelope}) bounds the difference between
$\hat s_\caus$ and $\slin$ under additional smoothness assumptions.

We next use a chosen similarity to define transport costs
and compare entire SAE dictionaries.

\section{Comparing SAE Dictionaries with Optimal Transport}\label{sec:stab}

This section introduces the energy weights that specify the mass assigned to each latent, and the
selected similarity that determines the pairwise cost. A transport plan
then gives the total discrepancy, its decomposition by latent, and
the fractions assigned to particular counterparts.

\subsection{Energy-Weighted Feature Distributions}

Fix an input distribution $\mathcal D$ and two trained SAEs.
For either SAE, define
\[
w_j:=\E_{\mathcal D}\|h_jd_j\|^2
=\E_{\mathcal D}h_j^2.
\]
We use $w_j$ as weights. The energy of the summed contributions is
\begin{align*}
\E_{\mathcal D}\Big\|\sum_j h_jd_j\Big\|^2
=\sum_j w_j+2\sum_{j<k}\E_{\mathcal D}[h_jh_k]\,
\langle d_j,d_k\rangle.
\end{align*}
\begin{definition}[Feature measures]\label{def:measures}
The energy measure of SAE $A$ assigns weight $w_j=\E_{\mathcal D}h_j^2$
to latent $j$. Its normalization is $a_j=w_j/\sum_\ell w_\ell$;
write $b_k$ for the corresponding weights of SAE $B$.
Given a unit embedding $\Phi$ from Lemma~\ref{lem:chordal}, the feature
distributions are
\[
\mu_A=\sum_j a_j\delta_{\Phi(j)},\qquad
\mu_B=\sum_k b_k\delta_{\Phi(k)}.
\]
\end{definition}

For activation geometry,
$\Phi(j)=h_j/\sqrt{\E_{\mathcal D}h_j^2}$ and
\begin{equation}
c_{jk}:=\|\Phi(j)-\Phi(k)\|^2
=2\left(1-\frac{\E[h_jh_k]}{\sqrt{\E h_j^2\E h_k^2}}\right).
\label{eq:uncentered-cost}
\end{equation}
All activation moments are uncentered. Nonnegative activations give
$0\le c_{jk}\le2$; the decoder and causal geometries have costs in
$[0,4]$. 


\subsection{The Transport Plan and Its Cost Decomposition}\label{ssec:plan}

A transport plan $P$ has the source and target weight vectors
$a$ and $b$ as its marginals: its row sums equal $a$, and its column sums
equal $b$. Let $\Pi(a,b)$ denote the set of such nonnegative matrices.
We select a plan by minimizing transport cost with an entropy penalty.
For fixed $\varepsilon>0$, define
\begin{equation}\label{def:cost}
P_\varepsilon=\pi_\varepsilon
:=\argmin_{P\in\Pi(a,b)}\;
\langle c,P\rangle+\varepsilon\KL(P\|a\otimes b).
\end{equation}
This plan is uniquely defined and is computed by Sinkhorn iteration
\citep{cuturi2013sinkhorn,peyre2019computational,nutz2021introduction}.
On positive marginal supports it has positive entries and depends
smoothly on costs and marginals.

\begin{definition}[Dictionary discrepancy and transported fractions]
\label{def:transport-readouts}
The dictionary-level transport cost is
\begin{equation}
D_\varepsilon(A,B;\mathcal D):=\langle c,P_\varepsilon\rangle.
\label{eq:transport-cost}
\end{equation}
For $a_j>0$, the fraction of latent $j$'s mass allocated to latent
$k$ is $T_{jk}=P_{\varepsilon,jk}/a_j$, and its mean transport cost is
$d_j^\mathrm{loc}=\sum_kT_{jk}c_{jk}$. For nonempty sets of source and
target indices $I,J$, write $a(I)=\sum_{j\in I}a_j$ and
$P_\varepsilon(I\times J)=\sum_{j\in I,k\in J}P_{\varepsilon,jk}$.
If $a(I)>0$, the fraction of mass from $I$ allocated to $J$ is
\[
\theta_{I,J}=\frac{P_\varepsilon(I\times J)}{a(I)}.
\]
\end{definition}
The marginal constraints and the definition of the cost give
\begin{gather}
\sum_k T_{jk}=1,\qquad \sum_j a_jT_{jk}=b_k,
\qquad T_{jk}\le\min\{1,b_k/a_j\},\label{eq:budget-readout}\\
D_\varepsilon=\sum_j a_jd_j^\mathrm{loc}
=\sum_{j,k}c_{jk}P_{\varepsilon,jk}.\label{eq:cost-decomposition}
\end{gather}
\emph{Equation~\eqref{eq:cost-decomposition} connects the two levels of
comparison: a latent contributes $a_jd_j^\mathrm{loc}$ to the
same cost used to compare the whole dictionaries. Summing over specified
latents locates their share of disagreement; the fractions $\theta_{I,J}$
describe how their mass is allocated across counterparts.}

\begin{definition}[Mass fractions satisfying a cost threshold]\label{def:plans}
For a cost threshold $\delta>0$, the fraction of latent $j$'s mass
allocated to partners with cost at most $\delta$ is
\[
\iden_\delta(j)=\frac1{a_j}\sum_{k:c_{jk}\le\delta}P_{\varepsilon,jk}.
\]
Its energy-weighted aggregate is $P_\varepsilon(\{c\le\delta\})$.
It describes the mass assigned to partners with similarity at least
$1-\delta/2$.
\end{definition}
Section~\ref{sec:fss} gives confidence intervals for fractions
$\theta_{I,J}$. 

\paragraph{Comparison across contexts.}
A context specifies an input distribution $\mathcal D_g$,
such as documents on a given topic. We keep the trained SAEs,
their retained indices, and $\varepsilon$ fixed across contexts,
computing weights, costs, and transport plans separately.

For $S\ge2$ SAEs, let $q=S(S-1)/2$ and define
\begin{equation}
M_g=\frac1q\sum_{s<t}
D_\varepsilon(A_s,A_t;\mathcal D_g),
\qquad
\Delta=M_g-M_h.
\label{eq:mean-seed-cost}
\end{equation}
We test $H_0:\Delta=0$ to compare mean costs across contexts.
For a fixed SAE pair and prespecified index sets $I,J$,
we also test
\[
H_{0,I,J}:\quad
\theta_{I,J}^{(g)}=\theta_{I,J}^{(h)}.
\]
For each SAE pair, Equation~\eqref{eq:cost-decomposition}
decomposes the cost difference into changes in latent contributions,
which may partially cancel.

\subsection{Models of Splitting and Absorption}
Two idealized models explain how transport fractions represent
splitting and absorption.  Full constructions are in Appendix~\ref{app:canonical}.
\begin{proposition}[Transport representation of a split]\label{prop:plan-split}
Let $h_j=z\ge0$ and $h_{k_i}=z\1_{A_i}$, where $(A_i)_{i=1}^\nu$
partition the input space and all latents share a unit decoder.
Put $m_i=\E[z^2\1_{A_i}]>0$ and $m=\sum_i m_i$.
On the block with marginals $\delta_j$ and
$\sum_i(m_i/m)\delta_{k_i}$, every feasible plan satisfies
$P(j,k_i)=m_i/m$. Moreover,
$\rho_\act(j,k_i)=\sqrt{m_i/m}$, so $\iden_\delta(j)=1$ for
$\delta\ge2-2\sqrt{\min_i m_i/m}$.
\end{proposition}
In the absorption model, one SAE represents a general feature and a
more specific feature separately. The second turns off the general
latent wherever the specific feature is active and uses a single latent
to represent both contributions there. The two original activations are
assumed proportional on that overlap.
\begin{proposition}[Transport in the absorption model]
\label{prop:plan-abs}
Consider the proportional-amplitude absorption model of
Appendix~\ref{app:canonical}, restricted to source latents
$(j_{\mathrm{gen}},j_{\mathrm{sp}})$ and target latents
$(k_{\mathrm{gen}},k_{\mathrm{abs}})$, in this order.
Use normalized activation energies as marginals $a,b$
and costs $c_{jk}=2(1-\rho_\act(j,k))$.

Let $\varphi\in(0,1)$ be the absorbed fraction of the
general latent's energy. The unregularized optimal plan
is unique and equals
\[
P_0=
\begin{pmatrix}
(1-\varphi)a_{j_{\mathrm{gen}}}
& \varphi a_{j_{\mathrm{gen}}}\\
0 & a_{j_{\mathrm{sp}}}
\end{pmatrix}.
\]
The entropic plans satisfy $P_\varepsilon\to P_0$
as $\varepsilon\downarrow0$.
\end{proposition}
These interpretations rely on the restricted
transport problems. In a full dictionary, all retained latents compete
for the same source and target mass. Appendix~\ref{app:det-stability}
bounds changes in the plan under cost perturbations and transport
between specified blocks, including blocks with unequal total masses.

\section{Statistical Inference for Transport Comparisons}\label{sec:fss}
Section~\ref{sec:stab} defines comparisons using population moments.
We now estimate them from a finite sample of documents. The same sample
determines activation costs and both marginals, so their estimation
errors must be considered jointly. We first give confidence intervals
for transported mass fractions, then for costs across several SAE pairs,
and finally for a difference between two contexts. The results condition
on the trained SAEs and independent selection data. They cover
activation or fixed decoder costs with activation-energy weights;
inference for causal costs requires a separate analysis.

\subsection{Document Sampling and Assumptions}

We assume a document $X$ to be randomly sampled. For its $L_X$ retained positions, two fixed SAEs provide activations
$h^{(1)}_j(X,t),h^{(2)}_k(X,t)\ge0$. The document contributes the following mean energies and cross-moments:
\begin{align*}
U_j(X):=L_X^{-1}\sum_t h^{(1)}_j(X,t)^2,\quad
V_k(X):=L_X^{-1}\sum_t h^{(2)}_k(X,t)^2,\quad
W_{jk}(X):=L_X^{-1}\sum_t h^{(1)}_j(X,t)h^{(2)}_k(X,t).
\end{align*}
Both SAEs process the same positions. Let $Y:=(U,V,\operatorname{vec}W)$, with
mean $\vartheta=\E Y=(u,v,\operatorname{vec}w)$. The mean energies define
the normalized weights, and the cross-moments define the activation
costs in \eqref{eq:uncentered-cost}. Replacing $\vartheta$ by
$\hat\vartheta=N^{-1}\sum_iY(X_i)$ gives the estimated weights, costs,
and plan $\hat P=P_\varepsilon(\hat\vartheta)$.

\begin{assumption}[Document moments]\label{ass:document-clt}
Conditional on the trained SAEs and independent selection information,
$X_1,\ldots,X_N$ are i.i.d., $\E\|Y(X)\|_2^2<\infty$, and all retained
energies $u_j,v_k$ are strictly positive. Dictionary sizes, the number
of reported functionals, and $\varepsilon>0$ are fixed as $N\to\infty$.
For several SAE pairs, the same conditions apply to their joint vector
of document moments, with a fixed finite number of SAEs.
\end{assumption}

Finite fourth moments of the activations, averaged over documents
and retained positions, suffice. Samples with
zero estimated energy for a retained latent are flagged as invalid for report. Appendix~\ref{app:moment-conventions} specifies
the definitions on these exceptional samples and the interpretation of selection and sampling from a fixed document pool.

\subsection{Confidence Intervals for Transported Mass Fractions}
First, we choose a finite list of nonempty source and target index sets
$(I_r,J_r)_{r=1}^F$ before evaluation, either a priori or on independent selection data. For each pair of sets, estimate the fraction of source mass transported to the target set:
\begin{equation}
\theta_r=\frac{P(I_r\times J_r)}{a(I_r)},\qquad
\hat\theta_r=\frac{\hat P(I_r\times J_r)}{\hat a(I_r)}.
\label{eq:family-fraction}
\end{equation}
The function below describes how a document's deviation from the population moments changes the estimated fractions to first order.
\begin{theorem}[Asymptotic inference for transported fractions]\label{thm:clt}
Under Assumption~\ref{ass:document-clt}, the map
$\mathcal T_\varepsilon(\vartheta)=(\theta_1,\ldots,\theta_F)$ is
continuously differentiable near the population moments. Put
$J_\varepsilon=D\mathcal T_\varepsilon(\vartheta)$,
$\psi(X)=J_\varepsilon(Y(X)-\vartheta)$, and
$\Sigma=\E[\psi\psi^\top]$. Then
\begin{align}
\hat{\boldsymbol\theta}-\boldsymbol\theta
&=N^{-1}\sum_i\psi(X_i)+o_p(N^{-1/2}),\label{eq:clt-linear}\\
\sqrt N(\hat{\boldsymbol\theta}-\boldsymbol\theta)
&\Rightarrow\mathcal N_F(0,\Sigma).\label{eq:clt-limit}
\end{align}
With $\hat\psi_i=D\mathcal T_\varepsilon(\hat\vartheta)
(Y(X_i)-\hat\vartheta)$, their sample covariance $\hat\Sigma$
converges in probability to $\Sigma$. Let $z_p$ denote the
$p$-quantile of the standard normal distribution. For $\Sigma_{rr}>0$, the pointwise
interval with $q=z_{1-\alpha/2}$ is
\begin{equation}
[\hat\theta_r-q\sqrt{\hat\Sigma_{rr}/N},\;
 \hat\theta_r+q\sqrt{\hat\Sigma_{rr}/N}]\cap[0,1],
\label{eq:clt-ci}
\end{equation}
and has asymptotic coverage $1-\alpha$. If all diagonal variances are
positive, choosing $q=z_{1-\alpha/(2F)}$ gives joint coverage with
$\liminf$ at least $1-\alpha$ for the specified list of fractions.
\end{theorem}
The variance includes uncertainty in the denominator, costs,
and both marginals. Appendix~\ref{app:clt-proof} derives the explicit
influence function \eqref{eq:clt-influence} and the covariance estimator.

\subsection{Joint Inference for Transport Costs}
The estimated transport costs for different SAE pairs use
the same document sample and can therefore be correlated.
We account for their covariance when constructing confidence
intervals for averages or differences of transport costs.

Fix $S\ge2$ trained SAEs, $q=S(S-1)/2$, and the pair set
$\mathcal P=\{(s,t):1\le s<t\le S\}$. Let $Y^{\rm all}(X)$ concatenate
each SAE's document energies and every pair's cross-moments. Every pair
uses the same $N$ documents. Write $\vartheta^{\rm all}=\E Y^{\rm all}$
and
\[
\mathbf D_\varepsilon
=\big(\langle c^{st},P^{st}_\varepsilon\rangle\big)_{(s,t)\in\mathcal P}.
\]
Let $\mathcal F:\vartheta^{\rm all}\mapsto\mathbf D_\varepsilon$.
As for the fractions, its derivative gives the vector of influence
functions:
\[
\boldsymbol\chi(X)=D\mathcal F(\vartheta^{\rm all})
\bigl(Y^{\rm all}(X)-\vartheta^{\rm all}\bigr).
\]
Appendix~\ref{app:cost-clt-proof} gives the explicit derivative and
formula \eqref{eq:cost-influence}, including variation in the cost and
both marginals.

\begin{theorem}[Joint CLT for transport costs]\label{thm:cost-clt}
Under Assumption~\ref{ass:document-clt} for $Y^{\rm all}$,
$\E\boldsymbol\chi=0$ and, with
$\Omega=\E[\boldsymbol\chi\boldsymbol\chi^\top]$,
\begin{align}
\widehat{\mathbf D}_\varepsilon-\mathbf D_\varepsilon
&=N^{-1}\sum_i\boldsymbol\chi(X_i)+o_p(N^{-1/2}),\label{eq:cost-linear}\\
\sqrt N(\widehat{\mathbf D}_\varepsilon-\mathbf D_\varepsilon)
&\Rightarrow\mathcal N_q(0,\Omega).\label{eq:cost-clt}
\end{align}
Let $\hat{\boldsymbol\chi}_i$ be the influence evaluated at the empirical
moments and plans, and let $\hat\Omega$ be its sample covariance.
Then $\hat\Omega\to_p\Omega$. For any fixed $v\in\R^q$ with
$v^\top\Omega v>0$,
\[
\frac{v^\top(\widehat{\mathbf D}_\varepsilon-\mathbf D_\varepsilon)}
{\sqrt{v^\top\hat\Omega v/N}}\Rightarrow\mathcal N(0,1).
\]
Consequently $v^\top\widehat{\mathbf D}_\varepsilon
\pm z_{1-\alpha/2}\sqrt{v^\top\hat\Omega v/N}$ is a pointwise
asymptotic $1-\alpha$ confidence interval.
\end{theorem}
The off-diagonal entries of $\Omega$ retain dependence between
SAE-pair estimates. Combining the influence functions for costs and
fractions also gives their joint covariance. Thus the same document
sample supports inference at both levels of comparison.

\paragraph{Document bootstrap.} Resampling whole documents with
the same sampled indices for every SAE pair preserves their dependence.
Recomputing weights, costs, and plans gives asymptotically valid
simultaneous intervals and an $\ell_1$ confidence radius for the plan (Appendix~\ref{ssec:bootstrap}).

\subsection{Comparing Context Distributions}
We apply the joint cost result to compare the same trained SAEs
across two topics. For topic $g$, with document distribution
$\mathcal D_g$, average the costs over all distinct SAE pairs \eqref{eq:mean-seed-cost}.

\begin{corollary}[A confidence interval for a context contrast]\label{cor:context-cost}
Let the two topic samples be independent, of sizes $N_g,N_h$, with the
same fixed SAEs and $\varepsilon$. Suppose the joint moment assumption
holds in each topic, $v^\top\Omega_gv,v^\top\Omega_hv>0$, and
$N_g/(N_g+N_h)\to\lambda\in(0,1)$. For $\Delta=M_g-M_h$, set
\[
\hat\Delta=\hat M_g-\hat M_h,\qquad
\widehat{\mathrm{se}}_\Delta^2
=\frac{v^\top\hat\Omega_gv}{N_g}+\frac{v^\top\hat\Omega_hv}{N_h}.
\]
Then $(\hat\Delta-\Delta)/\widehat{\mathrm{se}}_\Delta
\Rightarrow\mathcal N(0,1)$, and
$\hat\Delta\pm z_{1-\alpha/2}\widehat{\mathrm{se}}_\Delta$
has asymptotic coverage $1-\alpha$.
\end{corollary}
The interval concerns document sampling conditional on the trained SAEs; training-seed variation is assessed separately in Section~\ref{sec:exp}.

\section{Experiments on SAE Reproducibility}\label{sec:exp}

We compare SAEs trained with different initialization seeds, holding
model location, SAE width, and training data fixed. We use normalized
activation energies as weights and costs
$c_{jk}=2(1-\rho_{\mathrm{act}}(j,k))$, where
$\rho_{\mathrm{act}}$ is the normalized mean product of two latent
activations (Definition~\ref{def:kernels}).

\subsection{Toy Model: TinyStories-1M}

\paragraph{Model and data.}
We train five TopK SAEs ($k=16$, width 64, unit-norm decoders;
seeds 101--105) on the 64-dimensional output of block 3 (zero-based)
of the frozen \texttt{roneneldan/}\allowbreak\texttt{TinyStories-1M} model
\citep{eldan2023tinystories}. All five use the same training stories
and minibatch indices; FVU on the tuning data is 0.1425--0.1467.
Each topic, \texttt{animals} and \texttt{family}, contains 768 held-out
128-token prefixes with at least two words from its prespecified
list. Prefixes satisfying both lists are excluded. All SAEs are
evaluated at the same token positions. We assume that documents are
i.i.d.\ within each topic and that the two topic samples are independent.
Appendix~\ref{app:seed-protocol} describes hash assignment and selection
from a seeded document stream.

All 64 latents have positive energy and are active in at least
435 documents per topic. We compute ten pairwise transport plans
and five self-comparisons per topic. The main comparison uses
$\varepsilon=0.30$; the analyses of individual latents and larger
models use $\varepsilon=0.10$.

\paragraph{Difference between topics.}
The estimated mean transport cost is 1.02784 on \texttt{animals}
and 1.05424 on \texttt{family} (Figure~\ref{fig:context-sources}(a)).
The prespecified difference
$\Delta=M_{\rm family}-M_{\rm animals}$ is estimated as 0.02639,
or 2.57\% of the animals mean, with pointwise 95\% confidence interval
$[0.02277,0.03002]$. The interval accounts for document sampling and
dependence between SAE pairs evaluated on the same documents,
conditional on the trained SAEs
(Theorem~\ref{thm:cost-clt}; Corollary~\ref{cor:context-cost}).
All ten pairs have higher estimated cost on \texttt{family} at each
tested value of $\varepsilon$. Appendix~\ref{app:seed-protocol}
reports the full matrices and the costs of self-comparisons.

\paragraph{Variation across documents and training seeds.}
Repeating the computation on a second GPU changes the cost matrices
by at most $3\cdot10^{-4}$. The additional TinyStories-1M analyses
below use this second run. With 20 SAEs, all 190 pairs have positive estimated
differences at $\varepsilon\in\{0.10,0.30\}$. At $\varepsilon=0.30$,
the interval accounting for both document sampling and seed variation
excludes zero (Table~\ref{tab:seed-cost-main}). With five seeds, the
standard error due to seed variation is 0.0038, compared with 0.0018
for document sampling. These estimates of seed variation are separate
from the results conditional on trained SAEs in Section~\ref{sec:fss}.

\subsection{Decomposition and Individual Latents}

\paragraph{Changes in weights and costs.}
Both the energy weights and the cost matrix change between topics.
To separate their contributions, let $K_{uv}$ be the mean transport
cost over SAE pairs when both marginal distributions are taken from
topic $u$ and the cost matrix from topic $v$. Here $\mathrm{a}$ denotes
\texttt{animals} and $\mathrm{f}$ denotes \texttt{family}. We recompute
the optimal plan for each of the four combinations, using the same
$\varepsilon$. Thus $K_{\mathrm{aa}}=M_{\rm animals}$ and
$K_{\mathrm{ff}}=M_{\rm family}$. Define
\begin{align}
\Delta_w
&=\tfrac12\big[(K_{\mathrm{fa}}-K_{\mathrm{aa}})
              +(K_{\mathrm{ff}}-K_{\mathrm{af}})\big],\notag\\
\Delta_c
&=\tfrac12\big[(K_{\mathrm{af}}-K_{\mathrm{aa}})
              +(K_{\mathrm{ff}}-K_{\mathrm{fa}})\big].
\label{eq:exp-context-decomposition}
\end{align}
Then $\Delta=\Delta_w+\Delta_c$. Each term averages the change obtained
by replacing one quantity before or after the other.
In the second run with five SAEs at $\varepsilon=0.30$,
$\widehat\Delta_w=0.01680$ and $\widehat\Delta_c=0.00956$, which sum
to $\widehat\Delta=0.02636$. They account for 64\% and 36\% of the
estimated difference, respectively; both confidence intervals exclude
zero (Table~\ref{tab:seed-cost-main}). At $\varepsilon=0.10$, the
weight term accounts for 56\%.
Giving zero weight to topic-word positions, which make up 3.4\% of
all positions, leaves 78\% of the estimated difference at
$\varepsilon=0.30$.

\begin{table}[t]
\centering\small
\begin{tabular}{lrr}
\toprule
Quantity and source of variation & Estimate & 95\% interval\\
\midrule
$\Delta$, 5 seeds, documents & 0.02639 & $[0.02277,0.03002]$\\
$\Delta$, 5 seeds, jackknife & 0.02636 & $[0.01585,0.03687]$\\
$\Delta$, 20 seeds, documents and seeds & 0.02612 & $[0.02173,0.03050]$\\
\midrule
$\Delta_w$ (weights) & 0.01680 & $[0.01534,0.01826]$\\
$\Delta_c$ (cost matrix) & 0.00956 & $[0.00716,0.01195]$\\
\bottomrule
\end{tabular}
\caption{Estimates and confidence intervals for TinyStories-1M at
$\varepsilon=0.30$. The first row uses the original run; the others
use the second run. Means are over ten pairs for five seeds and
190 pairs for 20 seeds. Intervals for document sampling and for
$\Delta_w,\Delta_c$ condition on the trained SAEs. Seed variation is
estimated by a delete-one-seed jackknife, using the $t_4$ distribution
for five seeds. The 20-seed interval adds the document and seed
variance estimates.}
\label{tab:seed-cost-main}
\end{table}

\begin{figure}[!tbp]
\centering
\includegraphics[width=0.7\linewidth]{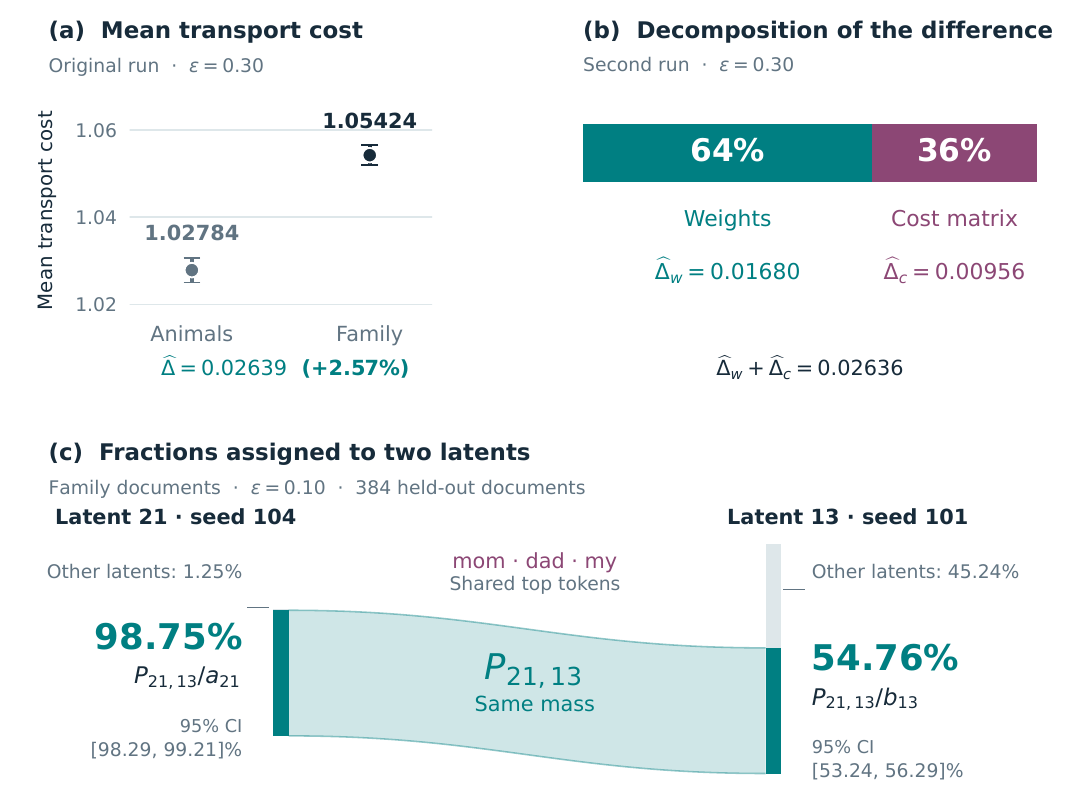}
\caption{Transport costs and fractions on TinyStories-1M.
(a) Mean costs over ten pairs of five SAEs in the original run
at $\varepsilon=0.30$, with pointwise 95\% confidence intervals
conditional on the trained SAEs.
(b) Decomposition of the difference $0.02636$ in the second run:
$0.01680$ (64\%) for weights and $0.00956$ (36\%) for the cost matrix.
We replace the animals weights and costs by the family values in
both orders, recompute the plan after each replacement, and average
the changes as in \eqref{eq:exp-context-decomposition}.
Percentages are fractions of the difference in mean costs.
(c) Family documents at $\varepsilon=0.10$.
Bar heights are proportional to $a_{21}$ and $b_{13}$; the band
represents their common plan entry $P_{21,13}$.
Gray portions are mass assigned to other latents.
Estimates and pointwise 95\% confidence intervals use 384 held-out
documents independent of selection. Proportions use rounded estimates.}
\label{fig:context-sources}
\label{fig:asymmetric-correspondence}
\label{fig:tinystories-panels}
\end{figure}

\paragraph{Uniform and energy weights.}
We compare entropic OT with uniform weights and with normalized
energy weights, keeping the activation cost matrices fixed.
Both allow mass from one latent to be assigned to several latents.
The bootstrap $z$-statistics are 1.5 and 15.0, respectively, with
positive values indicating greater discrepancy on \texttt{family}.
Unregularized OT with energy weights gives 15.4; hence the difference
is also present without entropy regularization. The Hungarian
algorithm gives 1.1, and the nearest-neighbour shared fraction gives
6.9 (Table~\ref{tab:app-baselines}). The reported comparisons based
only on decoder vectors do not depend on evaluation documents.
Since these methods estimate different quantities, their
$z$-statistics do not compare statistical power for the same hypothesis.

\begin{figure}[!tbp]
\centering
\includegraphics[width=0.7\linewidth]{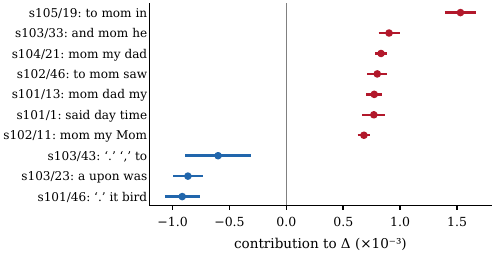}
\caption{Contributions of selected latents to $\Delta$ at
$\varepsilon=0.10$. Selection uses half of the documents; estimates
and 95\% confidence intervals use the other half. Labels give the
seed, latent index, and top tokens.}
\label{fig:latents}
\end{figure}

\paragraph{Contributions of individual latents.}
In the second run with five SAEs, we divide each term
$c_{jk}P_{\varepsilon,jk}$ in \eqref{eq:cost-decomposition} equally
between latents $j$ and $k$. Across the five SAEs, we select the ten latents
with the largest contributions to $\Delta$ on half of the documents
and estimate their contributions on the other half.
At $\varepsilon=0.10$, their contributions sum to 17\% of the
estimated difference. Latents associated with family words have
positive contributions; those associated with punctuation,
narrative, and animals have negative contributions
(Figure~\ref{fig:latents}).
The hypothesis of equal transported fractions between topics is
rejected in 91 of 120 tests after Bonferroni correction
(Theorem~\ref{thm:clt}).

\paragraph{Fractions assigned to a pair of latents.}
Latent 21 of seed 104 and latent 13 of seed 101 share top tokens
\texttt{mom}, \texttt{dad}, and \texttt{my}.
For these two SAEs, let $a_{21}$ and $b_{13}$ be their normalized
energy weights. On \texttt{family} documents at $\varepsilon=0.10$,
the estimated fractions are
\[
\frac{\widehat P_{21,13}}{\widehat a_{21}}=98.75\%,
\qquad
\frac{\widehat P_{21,13}}{\widehat b_{13}}=54.76\%.
\]
Thus almost all of latent 21's mass is assigned to latent 13, while
this same mass accounts for about 55\% of latent 13's total mass
(Figure~\ref{fig:asymmetric-correspondence}(c)). This example illustrates a graded notion of feature correspondence under the chosen activation geometry: latents with overlapping lexical profiles can account for different fractions of each other's normalized energy mass.

\subsection{Comparison Across Models and Widths}

We repeat the analysis on GPT-2 small, Pythia-160M, and TinyStories-33M,
with five seeds at each of two SAE widths, 3072 and 6144 ($d=768$).
Together with TinyStories-1M, this gives seven combinations of model
and width (Appendix~\ref{app:larger-models}).

\paragraph{Differences between topics.}
At $\varepsilon=0.10$, the estimated cost term $\Delta_c$ is positive
in GPT-2 and Pythia and near zero in TinyStories-33M. The estimated
weight term $\Delta_w$ takes both positive and negative values
across models and widths. In GPT-2 with five seeds and width 3072,
the positive cost term and negative weight term partially cancel
(Appendix~\ref{app:gpt2}). In a separate analysis with ten seeds
at width 3072, confidence intervals for $\Delta$ include zero in
all three models.

\paragraph{Transport costs for groups of latents.}
For each of the 35 SAEs, we estimate activation energies on independent
neutral tuning documents and sort the latents by decreasing energy.
We divide this ordered list into three groups containing approximately
50\%, 40\%, and 10\% of the total energy. For each group, we sum its
contributions to the transport costs at $\varepsilon=0.10$ using
the plans computed for the full dictionaries, then average over
the two topics.

\begin{figure}[!tbp]
\centering
\includegraphics[width=0.7\linewidth]{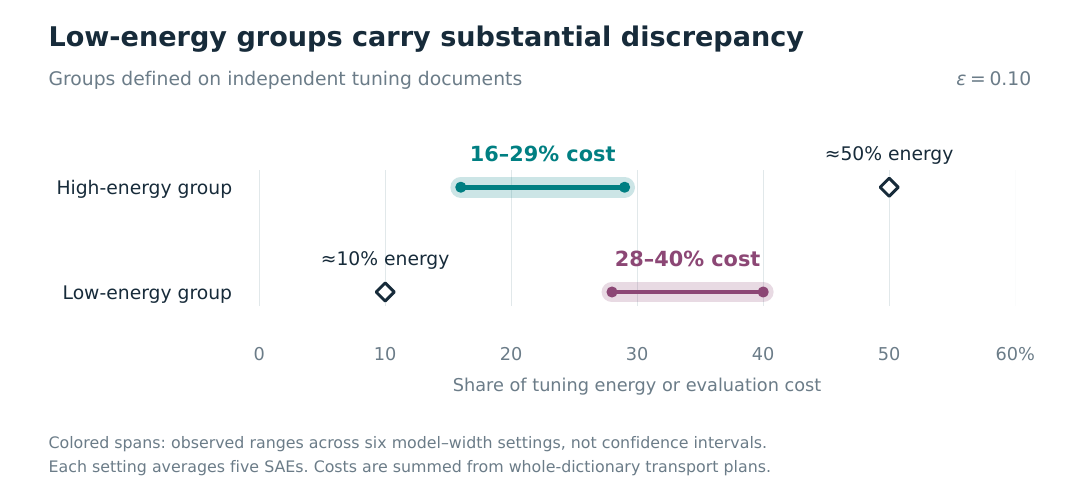}
\caption{Fractions of tuning energy and evaluation cost for groups
of latents at $\varepsilon=0.10$.
Diamonds show each group's approximate fraction of total tuning
energy. Colored segments show the minimum and maximum fractions
of evaluation cost across six combinations of model and width,
after averaging over five SAEs in each combination. These ranges
are not confidence intervals. Groups are defined using independent
neutral tuning documents. Cost contributions use the plans for
the full dictionaries and are averaged over both topics.}
\label{fig:energy-cost-groups}
\end{figure}

Across the six combinations with widths 3072 and 6144, the group
containing 50\% of tuning energy accounts for 16--29\% of evaluation
cost, whereas the group containing 10\% accounts for 28--40\%
(Figure~\ref{fig:energy-cost-groups}). Each value is an average over
five SAEs. Thus latents with low energy on the tuning data can
account for a substantial fraction of the transport cost, including
when the difference between topics is small.
We report these averages without confidence intervals
(Appendix~\ref{app:energy-groups}).

\subsection{Coverage and Regularization}

\paragraph{Coverage of confidence intervals.}
In a synthetic experiment with independent samples from a known
population, the empirical coverage of 95\% simultaneous
document-bootstrap intervals for three fractions and one cost is
94.5--95.5\%. In a separate experiment, rare documents with a large
effect on the estimates reduce coverage substantially below 95\%
(Appendix~\ref{ssec:calibration}). Resampling the fixed TinyStories
document pools gives 91.5--97.0\% coverage of 95\% intervals.
The latter experiment evaluates sampling from these pools
(Table~\ref{tab:app-coverage}).

\paragraph{Dependence on regularization.}
For the five TinyStories-1M SAEs, confidence intervals for $\Delta$
based on document sampling exclude zero and are positive at
$\varepsilon\in\{0.03,0.05,0.10,0.20,0.30,0.60\}$
(Table~\ref{tab:app-eps}). Correcting for the costs of comparing each
SAE with itself changes the estimated difference little at
$\varepsilon\le0.30$, but reduces it by about 75\% at
$\varepsilon=0.60$.

\section{Discussion}\label{sec:discussion}

A single transport plan connects the comparison of entire SAE
dictionaries with the analysis of individual latents. Its cost
decomposition identifies which latents contribute to the measured
difference, while transported fractions describe their allocation
across the other dictionary. Joint inference accounts for the fact
that costs, weights, and fractions are estimated from the same
documents.

The experiments show that comparisons based on activation patterns
depend on the evaluation distribution, even when the trained SAEs
are fixed. The positive difference between topics in TinyStories-1M
does not extend uniformly across models. Changes in weights and
cost matrices can partially cancel, so a small difference between
topic means does not imply that either term is small.
Moreover, groups with low activation energy on tuning documents
can account for a substantial fraction of evaluation cost.
These results support examining the decomposition alongside the
total cost.

Transported fractions require a similarly precise interpretation.
The fractions $P_{jk}/a_j$ and $P_{jk}/b_k$ describe the same mass
relative to different marginal weights and can therefore differ
substantially. They quantify correspondence under the specified
cost, input distribution, and regularization parameter.
They do not by themselves establish semantic identity or equal
effects on model outputs. The geometric results explain why
decoder, activation, and causal similarities can give different
comparisons.

The statistical guarantees concern document sampling conditional
on the trained SAEs, with positive retained energies and fixed
dictionary sizes and regularization. Training-seed variation is
assessed separately. The calibration experiments also show that
asymptotic validity does not ensure accurate coverage when rare
documents have a large effect on the estimates. Finally, the
sensitivity to regularization and self-comparison costs limits
interpretations that treat the reported values as independent
of these choices.

\bibliographystyle{plainnat}
\bibliography{references}

\clearpage
\onecolumn
\appendix

\section{Similarity Measures (Section~\ref{sec:setup})}

\subsection{Proofs for Section~\ref{sec:setup}}\label{app:proofs}

\paragraph{Proof of Lemma~\ref{lem:chordal}.}
Normalize $d_j$, $h_j$, and $e_j$ in $\R^n$, $L^2(\mathcal D)$,
and $L^2(\mathcal D;\R^V)$, respectively. The positive-energy and
nonzero-effect assumptions make these embeddings well defined, and their
inner products are the three similarities in
Definition~\ref{def:kernels}. For any two unit Hilbert-space vectors,
$\|\Phi(j)-\Phi(k)\|^2=2-2\langle\Phi(j),\Phi(k)\rangle=2-2s\in[0,4]$. \qed

\paragraph{Proof of Proposition~\ref{thm:t1uncon}.}
For a unit vector $g$, let $a=zg$, where $z\ge0$ has finite second
moment and positive probability in both $(0,2]$ and $(2,\infty)$.
Take $d_j=d_k=g$, $h_j=z\1\{z\le2\}$, and $h_k=z\1\{z>2\}$.
Then $h_jg+h_kg=a$, $s_\dec=1$, and the disjoint activation supports give
$\rho_\act=0$ with positive energies.
Conversely, take orthonormal $g_1,g_2$, $a=z(g_1+g_2)$,
$h_j=h_k=z$, $d_j=g_1$, and $d_k=g_2$, with $0<\E z^2<\infty$.
Moreover, $(s_\dec,\rho_\act)=(0,1)$. \qed

\paragraph{Proof of Proposition~\ref{thm:causal}.}
For constant $J$, the linearized effects $q_j=-h_jJd_j$ satisfy
$\E\langle q_j,q_k\rangle=\E[h_jh_k]\langle Jd_j,Jd_k\rangle$ and
$\E\|q_j\|^2=\E[h_j^2]\|Jd_j\|^2>0$.
Normalization with $M=J^\top J$ gives
$\slin=\rho_\act\cos_M(d_j,d_k)$, hence
$|\slin|\le\rho_\act$ by Cauchy--Schwarz.
For varying $J(x)$,
\[
\slin=\frac{\E\big[h_jh_k\langle J(x)d_j,J(x)d_k\rangle\big]}
{\sqrt{\E[h_j^2\|J(x)d_j\|^2]\,\E[h_k^2\|J(x)d_k\|^2]}}.
\]
Pointwise Cauchy--Schwarz and $h_j,h_k\ge0$ bound the absolute numerator
by $\E[h_jh_k\|J(x)d_j\|\|J(x)d_k\|]$, giving
$|\slin|\le\rho_J$. Cauchy--Schwarz in $L^2(\mathcal D)$ gives
$\rho_J\le1$; constant $J$ yields $\rho_J=\rho_\act$.
Finally, orthonormal $d_j,d_k$ and
$J=v(d_j+d_k)^\top/2$, $v\ne0$, give $Jd_j=Jd_k=v/2$.
Thus $s_\dec=0$ and $\slin=\rho_\act$, with the activation law free to
vary. \qed

\begin{proposition}[Negative linearized causal similarity with nearly parallel decoders]\label{prop:opp}
Fix orthonormal $\uu,\vv\in\R^n$ and $\epsilon\in(0,1)$, and set
$d_j=(\uu+\epsilon\vv)/\sqrt{1+\epsilon^2}$,
$d_k=(\uu-\epsilon\vv)/\sqrt{1+\epsilon^2}$, so
$s_\dec(j,k)=(1-\epsilon^2)/(1+\epsilon^2)\ge1-\epsilon$. If the (constant)
Jacobian $J$ of Assumption~\ref{ass:firstorder} annihilates the common
component, $J\uu=0$, $J\vv\ne0$, then $\cos_M(d_j,d_k)=-1$ and hence
$\slin(j,k)=-\rho_\act(j,k)<0$ whenever the latents co-activate:
nearly parallel decoders can therefore have opposite linearized effects.
\end{proposition}

\paragraph{Proof of Proposition~\ref{prop:opp}.}
Orthonormality gives unit decoders and
$s_\dec-(1-\epsilon)=\epsilon(1-\epsilon)^2/(1+\epsilon^2)\ge0$.
Since $J\uu=0$ and $J\vv\ne0$,
$Jd_j=-Jd_k=\epsilon J\vv/\sqrt{1+\epsilon^2}\ne0$.
Therefore $\cos_M=-1$, the effects have positive squared $L^2$ norms,
and Proposition~\ref{thm:causal} gives $\slin=-\rho_\act<0$ when
$\E[h_jh_k]>0$. \qed

\subsection{Second-Order Bound for Causal Similarity}\label{app:envelope}

\begin{proposition}[Bounds along the ablation path and after normalization]\label{prop:causal-path}
Let $d_j,d_k$ be unit decoders, $h_j,h_k\ge0$ have finite fourth moments,
and let $F$ be $C^2$ on an open set containing almost surely all segments
$\{a(X)-t h_\ell(X)d_\ell:0\le t\le1\}$ for $\ell=j,k$.
Assume finite deterministic $J_0,H_0$ satisfy almost surely
\[
\|DF(a(X))\|_{\rm op}\le J_0,\qquad
\sup_{\ell\in\{j,k\},\,0\le t\le1}
\|D^2F(a(X)-t h_\ell(X)d_\ell)\|\le H_0,
\]
where the Hessian norm is
$\sup_{\|u\|=\|v\|=1}\|D^2F(z)[u,v]\|$.
In $L^2(\mathcal D;\R^V)$ define
\[
q_\ell=-h_\ell Jd_\ell,\quad r_\ell=e_\ell-q_\ell,\quad
E_\ell=\|e_\ell\|_{L^2},\quad Q_\ell=\|q_\ell\|_{L^2},\quad
R_\ell=\tfrac12H_0\sqrt{\E h_\ell^4}.
\]
Suppose $E_j,E_k,Q_j,Q_k>0$. With
$s_e=\langle e_j,e_k\rangle/(E_jE_k)$ and
$s_q=\langle q_j,q_k\rangle/(Q_jQ_k)$,
\begin{align}
\|r_\ell(X)\|&\le\tfrac12H_0 h_\ell(X)^2,
&\|r_\ell\|_{L^2}&\le R_\ell,\nonumber\\
|s_e-s_q|&\le\min\{2,2R_j/E_j+2R_k/E_k\}.\label{eq:causal-envelope}
\end{align}
Moreover, with
$\rho_J=\E[h_jh_k\|Jd_j\|\|Jd_k\|]/(Q_jQ_k)$,
\[
|s_q|\le\rho_J\le1,\qquad
|s_e|\le\min\{1,\rho_J+2R_j/E_j+2R_k/E_k\}.
\]
For constant $J$ and nonzero directional effects, $\rho_J=\rho_\act$.
For a globally affine $F$, $H_0=0$ and $e=q$.
If $a,J,d_\ell,h_\ell$ are fixed and
$e_\ell(t)=F(a-t h_\ell d_\ell)-F(a)$ for $t\in(0,1]$, then whenever
$tR_\ell\le Q_\ell/2$ for both latents,
\[
\|e_\ell(t)\|_{L^2}\ge tQ_\ell/2,\qquad
|\cos(e_j(t),e_k(t))-s_q|
\le\min\{2,4t(R_j/Q_j+R_k/Q_k)\}.
\]
The $O(t)$ constant depends on the fixed positive first-order norms.
\end{proposition}
\begin{proof}
The integral Taylor formula on each full segment gives
\[
e_\ell=-h_\ell Jd_\ell+
 h_\ell^2\int_0^1(1-u)D^2F(a-u h_\ell d_\ell)[d_\ell,d_\ell]\nrev{\,du}.
\]
Unit decoder norm gives the pointwise remainder bound. Squaring and
integrating gives $R_\ell$; $\|q_\ell(X)\|\le J_0h_\ell(X)$ ensures all
L2 effects are finite. For nonzero Hilbert vectors $u,v$, adding and
subtracting $v/\|u\|$ and using the reverse triangle inequality yields
\[
\left\|\frac{u}{\|u\|}-\frac{v}{\|v\|}\right\|
\le\frac{\|u-v\|+|\|v\|-\|u\||}{\|u\|}
\le\frac{2\|u-v\|}{\|u\|}.
\]
The difference between inner products of two pairs of unit vectors is
bounded by the sum of the two vector differences. Apply this with
$u=e_\ell,v=q_\ell$ to prove \eqref{eq:causal-envelope}; two is the
universal cosine-difference bound. Pointwise Cauchy--Schwarz followed by
L2 Cauchy--Schwarz gives $|s_q|\le\rho_J\le1$.
Constant directional norms cancel from $\rho_J$; the affine case has zero
Hessian. Shortened segments give
$\|e_\ell(t)-tq_\ell\|_{L^2}\le t^2R_\ell$ and
$\|e_\ell(t)\|_{L^2}\ge tQ_\ell-t^2R_\ell\ge tQ_\ell/2$.
Substitution in the normalization inequality proves the final bound.
\end{proof}

Mixed activation moments also bound the numerator:
\[
|\langle e_j,e_k\rangle-\langle q_j,q_k\rangle|
\le\frac{J_0H_0}{2}\E[h_jh_k^2+h_j^2h_k]
  +\frac{H_0^2}{4}\E[h_j^2h_k^2],
\]
by expanding $e=q+r$; H\"older's inequality gives finiteness.
Equation~\eqref{eq:causal-envelope} additionally controls normalization.
The full-path Hessian assumption is essential: at $a=0,d=h=1$,
$F(z)=z+Kz^3$ has $F'(0)=1,F''(0)=0$ and ablation remainder $-K$.

\section{Optimal-Transport Comparison (Section~\ref{sec:stab})}

\subsection{Models of Feature Splitting and Absorption}\label{app:canonical}

\paragraph{Example 1: splitting.}
In the splitting model
\citep{bussmann2025matryoshka,leask2025canonical}, SAE $A$ has a parent
latent with decoder $d$ and activation $z\ge0$, $\E z^2>0$.
SAE $B$ replaces it by children with decoder $d$ and activations
$h_{k_i}=z\1_{A_i}$, where $\{A_i\}_{i=1}^{\nu}$ partitions the input
space. Their energies $m_i:=\E[z^2\1_{A_i}]$ sum to $m:=\E z^2$,
and their contributions sum to $zd$.

\paragraph{Example 2: absorption.}
In the absorption model \citep{chanin2024absorption}, a specific
feature implies a general one. SAE $A$ represents them separately;
SAE $B$ combines their contributions where they co-occur.

Let $d_{\mathrm{gen}}\perp d_{\mathrm{sp}}$ be unit decoder directions with
activations $h_{\mathrm{gen}},h_{\mathrm{sp}}\ge0$: $d_{\mathrm{gen}}$ carries
the general feature on a set of tokens $G$, and $d_{\mathrm{sp}}$ the specific
one on $S\subseteq G$ (the specific feature implies the general). On $S$ the
two are proportional, $h_{\mathrm{gen}}=c\,h_{\mathrm{sp}}$ with $c>0$, and
$\Pr(S)>0$, $\Pr(G\setminus S)>0$. SAE $A$ encodes them as separate
latents $j_{\mathrm{gen}}=(d_{\mathrm{gen}},h_{\mathrm{gen}})$ and
$j_{\mathrm{sp}}=(d_{\mathrm{sp}},h_{\mathrm{sp}})$. SAE $B$ retains
the general feature only outside $S$ and replaces the overlap by a single
latent with a different decoder direction,
\begin{gather*}
h_{k_{\mathrm{gen}}}=h_{\mathrm{gen}}\1_{G\setminus S},\qquad
h_{k_{\mathrm{abs}}}=\sqrt{1+c^2}\,h_{\mathrm{sp}},\qquad
d_{\mathrm{abs}}=\frac{d_{\mathrm{sp}}+c\,d_{\mathrm{gen}}}{\sqrt{1+c^2}};
\end{gather*}
The combined contribution on $S$ is unchanged. Write
$w_g:=\E[h_{\mathrm{gen}}^2]$ and
$\varphi:=\E[h_{\mathrm{gen}}^2\1_S]/w_g\in(0,1)$ for the absorbed energy fraction. With rows $(j_{\rm gen},j_{\rm sp})$ and columns
$(k_{\rm gen},k_{\rm abs})$, the activation similarities are
\[
\rho_\act=
\begin{pmatrix}\sqrt{1-\varphi}&\sqrt\varphi\\0&1\end{pmatrix}.
\]
These entries follow from disjoint supports and proportionality
on $S$. Proposition~\ref{prop:plan-abs} identifies $\varphi$ with a
transport fraction in this model as $\varepsilon\to0$.

\subsection{Stability and Transport Between Blocks}\label{app:det-stability}
This section bounds how transport plans change when the costs are
perturbed while the marginals remain fixed. It also bounds transport
between specified blocks under a separation assumption on the costs.

\begin{definition}[Total variation between transport plans]\label{def:disagree}
Consider two similarity notions with costs $c^{(1)},c^{(2)}$ computed on the supports of
$\mu_A$ and $\mu_B$, whose plans are built on common marginals (e.g.\ the
activation-energy marginals). Define
$\Dis:=\tfrac12\|\pi^{(1)}_\varepsilon-\pi^{(2)}_\varepsilon\|_1\in[0,1]$, the
total-variation distance between the two transport plans on the product
of the latent index sets.

\end{definition}

\begin{theorem}[Cost-perturbation stability]\label{thm:stab}
Fix $\varepsilon>0$ and marginals $\mu_A,\mu_B$, let $\pi_\varepsilon$ and $\hat\pi_\varepsilon$ be the
entropic plans \eqref{def:cost} for costs $c$ and $\hat c$, and set
$\delta:=\|\hat c-c\|_\infty$. Then,
\begin{gather*}
\KL(\hat\pi_\varepsilon\,\|\,\pi_\varepsilon)\le\frac{2\delta^2}{\varepsilon^2},\qquad
\|\hat\pi_\varepsilon-\pi_\varepsilon\|_1\le\frac{2\delta}{\varepsilon},\qquad
\big|\hat\pi_\varepsilon(Q)-\pi_\varepsilon(Q)\big|\le\frac{\delta}{\varepsilon}\ \ \text{for every fixed cell set }Q.
\end{gather*}
\end{theorem}
\begin{proof}
Set $R=\mu_A\otimes\mu_B$ and
$F_c(\pi')=\langle c,\pi'\rangle+\varepsilon\KL(\pi'\|R)$.
The optimality equations give
$\pi=e^{(f\oplus g-c)/\varepsilon}R$
\citep{nutz2021introduction}. For feasible $\pi'$, the potential term
$\sum\pi'(f\oplus g)=\langle f,\mu_A\rangle+\langle g,\mu_B\rangle$
is constant, yielding
$F_c(\pi')-F_c(\pi)=\varepsilon\KL(\pi'\|\pi)$
\citep{csiszar1975idivergence}. Optimality of $\hat\pi$ for $\hat c$ then gives
\[
\varepsilon\KL(\hat\pi\|\pi)=F_c(\hat\pi)-F_c(\pi)
=\underbrace{F_{\hat c}(\hat\pi)-F_{\hat c}(\pi)}_{\le0}
+\langle c-\hat c,\hat\pi-\pi\rangle
\le\delta\,\|\hat\pi-\pi\|_1
\]
by H\"older's inequality. Pinsker's inequality
$\|\hat\pi-\pi\|_1\le\sqrt{2\KL(\hat\pi\|\pi)}$ yields
$\KL(\hat\pi\|\pi)\le2\delta^2/\varepsilon^2$ and
$\|\hat\pi-\pi\|_1\le2\delta/\varepsilon$.
Total variation gives the final bound
$|\hat\pi(Q)-\pi(Q)|\le\|\hat\pi-\pi\|_1/2\le\delta/\varepsilon$. 
\end{proof}

\begin{remark}\label{rem:eps-necessary}
Theorem~\ref{thm:stab} quantifies worst-case cost sensitivity as a
function of $\varepsilon$. Proposition~\ref{prop:tight} exhibits the
growth of this sensitivity at small regularization.
\end{remark}

The next result concerns a specified partition of the latent indices into paired blocks. The full transport marginals remain fixed, but corresponding blocks may have unequal total mass. Their imbalance determines how much
transport between different blocks is unavoidable.

\begin{theorem}[Transport between blocks with unequal masses]\label{thm:block}
Let $\mu_A,\mu_B$ be probability marginals on finite latent sets,
partitioned into paired blocks $(A_r,B_r)_{r=1}^K$, with $K\ge1$.
Write
\[
\eta_{\mathrm{blk}}
:=\frac12\sum_{r=1}^K|\mu_A(A_r)-\mu_B(B_r)|,\qquad
Q_{\mathrm{cross}}:=\bigcup_{r\ne s}A_r\times B_s.
\]
Assume $0\le c\le\bar c$, with $c\le c_{\mathrm{in}}$ on
$\bigcup_r A_r\times B_r$ and $c\ge c_{\mathrm{out}}$ on
$Q_{\mathrm{cross}}$, where
$0\le c_{\mathrm{in}}<c_{\mathrm{out}}\le\bar c$.
Set $\gamma:=c_{\mathrm{out}}-c_{\mathrm{in}}>0$.
For the entropic plan \eqref{def:cost} at $\varepsilon>0$, let
$L_\varepsilon:=\pi_\varepsilon(Q_{\mathrm{cross}})$. Then
\begin{equation}
\eta_{\mathrm{blk}}\le L_\varepsilon\le
\min\left\{1,\,
\eta_{\mathrm{blk}}+
\frac{(\bar c-c_{\mathrm{out}})\eta_{\mathrm{blk}}
     +\varepsilon\log(K+1)}{\gamma}\right\}.
\label{eq:block-leakage-imbalance}
\end{equation}
More precisely, with
$h_2(t):=-t\log t-(1-t)\log(1-t)$ and $0\log0:=0$,
\begin{equation}
\gamma(L_\varepsilon-\eta_{\mathrm{blk}})
\le(\bar c-c_{\mathrm{out}})\eta_{\mathrm{blk}}
 +\varepsilon\{h_2(L_\varepsilon)+L_\varepsilon\log K\}.
\label{eq:block-leakage-entropy}
\end{equation}
The lower bound $\eta_{\mathrm{blk}}$ is attainable over all
feasible couplings.
\end{theorem}
\begin{proof}
Write $P=\pi_\varepsilon$, $L=P(Q_{\mathrm{cross}})$,
$\eta=\eta_{\mathrm{blk}}$, and
$a_r=\mu_A(A_r)$, $b_r=\mu_B(B_r)$.
Every coupling places at most $\min(a_r,b_r)$ on $A_r\times B_r$, so
\[
L\ge1-\sum_r\min(a_r,b_r)
=\frac12\sum_r|a_r-b_r|=\eta.
\]
The claims are immediate for $L=0$. Below take $L>0$ and omit terms
with weight $1-L$ when $L=1$.

\emph{A coupling attaining $\eta$.}
Normalize the within-block and cross-block restrictions of $P$:
$P=(1-L)P_{\mathrm{in}}+LP_{\mathrm{out}}$.
Let $\alpha,\beta$ be the marginals of $P_{\mathrm{out}}$ and
$d_r=P(A_r\times B_r)$. Their block masses satisfy
\[
u_r=\alpha(A_r)=\frac{a_r-d_r}{L},\qquad
v_r=\beta(B_r)=\frac{b_r-d_r}{L},\qquad
d:=\frac12\sum_r|u_r-v_r|=\frac{\eta}{L}.
\]
For $d>0$, couple $u,v$ by
\[
T_{rs}=\1_{\{r=s\}}\min(u_r,v_r)
       +\frac{(u_r-v_r)_+(v_s-u_s)_+}{d},
\]
and for $d=0$ take $T=\operatorname{diag}(u)$. This has marginals $u,v$
and off-diagonal mass $d$. Define
\[
\widetilde P_{\mathrm{out}}(j,k)
=T_{rs}\frac{\alpha_j}{u_r}\frac{\beta_k}{v_s},
\qquad j\in A_r,\quad k\in B_s,
\]
omitting blocks with zero residual marginal.
Its marginals are $\alpha,\beta$, so
\[
Q=(1-L)P_{\mathrm{in}}+L\widetilde P_{\mathrm{out}}
\]
has the required marginals and cross-block mass $Ld=\eta$.

\emph{Cost and entropy changes.}
Replacing $P$ by $Q$ moves $L-\eta$ mass inside blocks and leaves
$\eta$ outside, giving
\[
\langle c,Q-P\rangle
\le c_{\mathrm{in}}(L-\eta)+\bar c\,\eta-c_{\mathrm{out}}L
=-\gamma(L-\eta)+(\bar c-c_{\mathrm{out}})\eta.
\]

Write $H$ for Shannon entropy of a finite probability table or
vector, using natural logarithms. Since $P,Q$ have equal marginals,
\[
\KL(Q\|\mu_A\otimes\mu_B)-\KL(P\|\mu_A\otimes\mu_B)
=H(P)-H(Q).
\]
Disjoint supports for the restrictions of $P$ and concavity for $Q$ give
\begin{align*}
H(P)=h_2(L)+(1-L)H(P_{\mathrm{in}})+LH(P_{\mathrm{out}}),\quad
H(Q)\ge(1-L)H(P_{\mathrm{in}})
                  +LH(\widetilde P_{\mathrm{out}}).
\end{align*}
The residual plans share marginals $\alpha,\beta$, hence
\begin{align*}
H(P_{\mathrm{out}})-H(\widetilde P_{\mathrm{out}})
&=\KL(\widetilde P_{\mathrm{out}}\|\alpha\otimes\beta)
  -\KL(P_{\mathrm{out}}\|\alpha\otimes\beta)\\
&\le\KL(\widetilde P_{\mathrm{out}}\|\alpha\otimes\beta)\\
&=\KL(T\|u\otimes v)
\le\min\{H(u),H(v)\}\le\log K.
\end{align*}
Here conditional independence within each block gives the equality;
the remaining bound is the entropy bound on mutual information.
Thus
\[
\KL(Q\|\mu_A\otimes\mu_B)-\KL(P\|\mu_A\otimes\mu_B)
\le h_2(L)+L\log K\le\log(K+1).
\]
The maximum of $h_2(t)+t\log K$ on $[0,1]$ is $\log(K+1)$.

Optimality of $P$ gives
\[
\langle c,P-Q\rangle
\le\varepsilon\{
\KL(Q\|\mu_A\otimes\mu_B)-\KL(P\|\mu_A\otimes\mu_B)\}.
\]
Substitution of the cost and entropy bounds proves
\eqref{eq:block-leakage-entropy}; maximizing the entropy bound and
using $L\le1$ gives \eqref{eq:block-leakage-imbalance}.

\emph{Balanced comparison bound.}
If $a_r=b_r=m_r$, the block-diagonal product plan
$\pi_{\mathrm{bl}}(j,k)=\mu_A(j)\mu_B(k)/m_r$ on
$A_r\times B_r$ is feasible; zero-mass blocks are omitted.
It has cost at most $c_{\mathrm{in}}$ and
$\KL(\pi_{\mathrm{bl}}\|\mu_A\otimes\mu_B)=H_{\mathrm{blk}}$.
Optimality and nonnegative costs give
\[
c_{\mathrm{out}}L
\le\langle c,P\rangle
\le c_{\mathrm{in}}+\varepsilon H_{\mathrm{blk}}.
\]
This proves the balanced-case bound. 
\end{proof}
The partition and cost gap are specified assumptions.
The term $(\bar c-c_{\mathrm{out}})\eta_{\mathrm{blk}}$ accounts for
variation in the cost of reallocating the unavoidable cross-block mass.
One may take $\bar c=2$ for nonnegative activation geometry and
$\bar c=4$ for decoder or causal geometry.

\paragraph{Balanced special case.}

If $\mu_A(A_r)=\mu_B(B_r)=m_r$ for every $r$, then
$\eta_{\mathrm{blk}}=0$. Comparison with a block-diagonal coupling that
has product distributions within each paired block also gives a bound
in terms of the entropy of the block masses. Combining the two bounds
gives
\[
L_\varepsilon\le
\min\left\{1,\,
\frac{c_{\mathrm{in}}+\varepsilon H_{\mathrm{blk}}}
     {c_{\mathrm{out}}},\,
\frac{\varepsilon\log(K+1)}{\gamma}\right\},
\qquad
H_{\mathrm{blk}}:=\sum_r m_r\log(1/m_r).
\]
The smaller bound depends on the parameters; the cost-gap bound
tends to zero as $\varepsilon\to0$ for fixed $\gamma>0$.

\begin{corollary}[Transport between blocks under cost perturbations]
\label{cor:epsstar}
Under the hypotheses of Theorem~\ref{thm:block}, keep the same
marginals and block partition for the population and perturbed
problems, and suppose a supplied bound satisfies
$\|\hat c-c\|_\infty\le\delta$. Then
\[
\hat\pi_\varepsilon(Q_{\mathrm{cross}})
\le\min\left\{1,\,
\eta_{\mathrm{blk}}+
\frac{(\bar c-c_{\mathrm{out}})\eta_{\mathrm{blk}}
     +\varepsilon\log(K+1)}{\gamma}
+\frac{\delta}{\varepsilon}\right\}.
\]
For $\delta>0$, the second argument of the minimum is minimized at $\varepsilon^\star=
\sqrt{\frac{\delta\gamma}{\log(K+1)}},$
where its value is
$\eta_{\mathrm{blk}}+
\frac{(\bar c-c_{\mathrm{out}})\eta_{\mathrm{blk}}}{\gamma}
+2\sqrt{\frac{\delta\log(K+1)}{\gamma}}.$
\end{corollary}
\begin{proof}
Apply Theorem~\ref{thm:stab} to the fixed set
$Q_{\mathrm{cross}}$ to obtain
$\hat\pi_\varepsilon(Q_{\mathrm{cross}})
\le\pi_\varepsilon(Q_{\mathrm{cross}})+\delta/\varepsilon$.
Combining this with \eqref{eq:block-leakage-imbalance} and
$\hat\pi_\varepsilon(Q_{\mathrm{cross}})\le1$ proves the first claim.
The only $\varepsilon$-dependent terms before clipping are
$\varepsilon\log(K+1)/\gamma+\delta/\varepsilon$.
For $\delta>0$, their minimum is attained at
$\varepsilon^\star=\sqrt{\delta\gamma/\log(K+1)}$,
where their sum is $2\sqrt{\delta\log(K+1)/\gamma}$. 
\end{proof}

\subsection{Proofs for Section~\ref{sec:stab}}\label{app:stab-proof}

\paragraph{Standing hypotheses for the transport arguments.}
Finite supports and costs give a compact, nonempty coupling
polytope and a unique positive entropic optimum on the positive marginal supports for $\varepsilon>0$. Statistical arguments condition on the
trained SAEs and independent selection data, using Assumption~\ref{ass:document-clt} for the CLTs and the moment event in
Appendix~\ref{app:fss} for finite-sample bounds.

\paragraph{Proof of Proposition~\ref{prop:plan-split}.}
The column marginals force $\pi(j,k_i)=m_i/m$, so the feasible set is a singleton for both unregularized and entropic transport. Since $\E[h_jh_{k_i}]=m_i$, $\E h_j^2=m$, and $\E h_{k_i}^2=m_i$, the cost is $c(j,k_i)=2-2\sqrt{m_i/m}$. For $\delta\ge2-2\sqrt{\min_i m_i/m}$, every cell satisfies $c\le\delta$ and $\iden_\delta(j)=\sum_i m_i/m=1$. \qed

\paragraph{Derivation of the absorption model.}\label{app:absorption-model}
Under Assumption~\ref{ass:abs}, orthogonality makes
$d_{\mathrm{abs}}=(d_{\mathrm{sp}}+c\,d_{\mathrm{gen}})/\sqrt{1+c^2}$
a unit vector. On $S$,
$h_{\mathrm{gen}}d_{\mathrm{gen}}+h_{\mathrm{sp}}d_{\mathrm{sp}}
=\sqrt{1+c^2}\,h_{\mathrm{sp}}d_{\mathrm{abs}}$;
off $S$, both representations retain $h_{\mathrm{gen}}d_{\mathrm{gen}}$.
Thus the construction in Appendix~\ref{app:canonical} preserves the
reconstruction with one active latent on $S$.

\paragraph{Proof of Proposition~\ref{prop:plan-abs}.}
Write $w_s=\E h_{\mathrm{sp}}^2$. The unnormalized marginals are
$(w_g,w_s)$ and $((1-\varphi)w_g,(1+c^2)w_s)$; they have equal total
mass because $\varphi w_g=c^2w_s$. With the row and column order of
Appendix~\ref{app:canonical}, every feasible plan has the form
\[
\pi(t)=\begin{pmatrix}
(1-\varphi)w_g-t&\varphi w_g+t\\
t&w_s-t
\end{pmatrix},\qquad
0\le t\le\min\{(1-\varphi)w_g,w_s\}.
\]
The activation costs in that appendix give cost slope
$2-2\sqrt\varphi+2\sqrt{1-\varphi}>0$ for $\varphi\in(0,1)$.
Thus the unique optimum is $\pi(0)$, whose general-to-absorbed fraction
equals $\varphi$. Normalization rescales all entries and preserves this
ratio. The entropy term is continuous and bounded on the compact
coupling polytope, so the entropic objectives converge uniformly to the
linear cost as $\varepsilon\to0$. Every cluster point therefore equals
the unique optimum $\pi(0)$. \qed

\paragraph{Absorption under causal costs.}
Take $c=2$ and an affine scalar output map with $Jd_{\rm gen}=1$, $Jd_{\rm sp}=-1$. Then $Jd_{\rm abs}=1/\sqrt5>0$ and the causal similarity matrix, with rows (general, specific) and columns (remaining, absorbed), is
\[
\begin{pmatrix}\sqrt{1-\varphi}&\sqrt\varphi\\0&-1\end{pmatrix}.
\]
The cost slope is $2[\sqrt{1-\varphi}-\sqrt\varphi-1]<0$, so the opposite endpoint of the feasible interval is optimal. Normalization preserves this sign. The construction is realized by $w_s=\varphi w_g/4$ and positive amplitudes on $S$ and $G\setminus S$; it shows how the
interpretation of the fraction depends on the cost.
\begin{proposition}[Tightness]\label{prop:tight}
For every $\varepsilon>0$ and $\delta\in(0,2\varepsilon]$ there are two-latent
instances with $\|\hat c-c\|_\infty=\delta$ and
$\|\hat\pi-\pi\|_1=\tanh\!\big(\delta/(2\varepsilon)\big)$, which matches the
$2\delta/\varepsilon$ bound of Theorem~\ref{thm:stab} up to an absolute
constant if $\delta\le2\varepsilon$.
\end{proposition}

\paragraph{Proof of Proposition~\ref{prop:tight}.}
Take $\mu_A=\mu_B=(\tfrac12,\tfrac12)$, $c=0$, and
$\hat c$ equal to zero on the diagonal and $\delta$ off it.
Then $\pi=\frac14\mathbf1$, while symmetry and row normalization give
$\hat\pi=\frac1{2(1+q)}\left(\begin{smallmatrix}1&q\\q&1\end{smallmatrix}\right)$,
$q=e^{-\delta/\varepsilon}$. Hence
$\|\hat\pi-\pi\|_1=(1-q)/(1+q)=\tanh(\delta/(2\varepsilon))$, which is at least $\tanh(1)\delta/(2\varepsilon)$ for $\delta\le2\varepsilon$. \qed

\begin{lemma}[Bounds for mass fractions satisfying a cost threshold]\label{lem:sandwich}
Let $P\in\Pi(a,b)$ and $\hat P\in\Pi(\hat a,\hat b)$, with
$a_j,\hat a_j>0$, and suppose supplied bounds satisfy
\[
\|\hat c-c\|_\infty\le\delta_c,\quad
\sum_k|\hat P_{jk}-P_{jk}|\le E_j,\quad
|\hat a_j-a_j|\le D_j,\quad a_j\ge\ell_j>0.
\]
For $\hat f_j(t)=\hat a_j^{-1}\sum_{k:\hat c_{jk}\le t}\hat P_{jk}$,
the population thresholded fraction satisfies
\[
\max\left\{0,\hat f_j(\delta-\delta_c)-\frac{E_j+D_j}{\ell_j}\right\}
\le\iden_\delta(j)\le
\min\left\{1,\hat f_j(\delta+\delta_c)+\frac{E_j+D_j}{\ell_j}\right\}.
\]
One computable choice is $\ell_j=\hat a_j-D_j$ when positive.
If $\|\hat P-P\|_1\le\Delta$, the aggregate mass with cost at most $\delta$ is bounded
by $\hat P(\{\hat c\le\delta-\delta_c\})-\Delta/2$ and
$\hat P(\{\hat c\le\delta+\delta_c\})+\Delta/2$, clipped to $[0,1]$.
\end{lemma}

\begin{proof}
Cost closeness implies
$\{\hat c\le\delta-\delta_c\}\subseteq\{c\le\delta\}
\subseteq\{\hat c\le\delta+\delta_c\}$.
Replacing a subset sum in row $j$ of $P$ by the corresponding sum in
$\hat P$ changes it by at most $E_j$. For any subset sum $S\le\hat a_j$,
\[
\left|\frac{S}{a_j}-\frac{S}{\hat a_j}\right|
=\frac{S}{\hat a_j}\frac{|\hat a_j-a_j|}{a_j}
\le D_j/\ell_j.
\]
Dividing the row error by $a_j\ge\ell_j$ proves the first pair of bounds. For an aggregate set $Q$, two probability tables satisfy $|\hat P(Q)-P(Q)|\le\|\hat P-P\|_1/2$; apply this to the two threshold sets.
\end{proof}

\subsection{Composition, Correlation Correction, and Unbalanced Transport}

\begin{proposition}[Composition under nested partitions of activation supports]\label{prop:comp}
Consider two successive partitions of activation supports, denoted
$A\to B\to C$. Parent activations
$z_j\ge0$ have disjoint supports and masses
$m^{(j)}=\E z_j^2>0$, $\sum_jm^{(j)}=1$. Partition each support into
child sets $A_{j,i}$ and each child into grandchild sets $B_{j,i,t}$,
with activations $z_j\1_{A_{j,i}}$ and $z_j\1_{B_{j,i,t}}$.
All child masses $m_{j,i}>0$ and grandchild masses $m_{j,i,t}>0$ are fixed,
$\sum_i m_{j,i}=m^{(j)}$, $\sum_t m_{j,i,t}=m_{j,i}$.
Use activation cost $2(1-\rho_\act)$ at all three stages and the corresponding
energy marginals. Define
\[
\pi_{AC}^{\rm comp}=\pi_{AB}\operatorname{diag}(\mu_B)^{-1}\pi_{BC},\qquad
R_\varepsilon=\tfrac12\|\pi_{AC}^{\rm direct}-\pi_{AC}^{\rm comp}\|_1.
\]
At $\varepsilon=0$, all three optima are unique and their only nonzero
entries are
\[
\pi_{AB}^0(j,k_{j,i})=m_{j,i},\quad
\pi_{BC}^0(k_{j,i},l_{j,i,t})=m_{j,i,t},\quad
\pi_{AC}^0(j,l_{j,i,t})=m_{j,i,t}.
\]
Thus $R_0=0$. For these activation costs and fixed positive masses,
$R_\varepsilon=O(\varepsilon)$ as $\varepsilon\downarrow0$.
\end{proposition}

\paragraph{Proof of Proposition~\ref{prop:comp}.}

Each target $k$ has one ancestor $f(k)$ with cost
$c_{*,k}=2(1-\sqrt{m_k/m_{f(k)}})<2$; disjoint supports give cost two
elsewhere. Ancestors are parents for $A\to B,A\to C$ and children for
$B\to C$. Assigning each column's mass to its ancestor is feasible, and
for any other feasible $P$,
\[
\langle c,P\rangle-\langle c,P^0\rangle
=\sum_k\sum_{r\ne f(k)}(2-c_{*,k})P_{rk}.
\]
Positive coefficients prove uniqueness. The Markov product telescopes to
$m_{j,i}m_{j,i,t}/m_{j,i}=m_{j,i,t}$, proving $R_0=0$.

Since $\|P-P^0\|_1=2\sum_k\sum_{r\ne f(k)}P_{rk}$, the displayed identity
gives $\langle c,P\rangle-\langle c,P^0\rangle\ge\sigma\|P-P^0\|_1$
for $\sigma=\tfrac12\min_k(2-c_{*,k})>0$.
With $R=a\otimes b$, entropic optimality gives
\[
\sigma\|P_\varepsilon-P^0\|_1
\le\varepsilon\{\KL(P^0\|R)-\KL(P_\varepsilon\|R)\}
\le\varepsilon\KL(P^0\|R).
\]
Thus each plan differs from its limit by $O(\varepsilon)$. The map
$(U,V)\mapsto U\operatorname{diag}(\mu_B)^{-1}V$ is locally Lipschitz
for fixed positive intermediate masses, giving $R_\varepsilon=O(\varepsilon)$.
\qed

\begin{proposition}[Correction for attenuation of correlations]\label{prop:atten}
Let $S_1,S_2$ be population statistics across latents
and $\hat S_i=S_i+N_i$ their estimates, with mean-zero noises $N_i$
independent of $(S_1,S_2)$ and of each other, and reliabilities
$r_i:=\Var(S_i)/\Var(\hat S_i)\in(0,1]$. Then
$\corr(\hat S_1,\hat S_2)=\sqrt{r_1r_2}\;\corr(S_1,S_2)$; the reliability
$r_i$ equals the correlation of two independent replicates
$\hat S_i^{(1)},\hat S_i^{(2)}$ of the same statistic; and the disattenuated
estimator $\corr(\hat S_1,\hat S_2)/\sqrt{\hat r_1\hat r_2}$, with $\hat r_i$
the replicate correlations, is consistent for $\corr(S_1,S_2)$ provided the empirical
correlations consistently estimate their population
counterparts \citep{spearman1904proof}.
\end{proposition}

\paragraph{Proof of Proposition~\ref{prop:atten}.}
Independence gives $\Cov(\hat S_1,\hat S_2)=\Cov(S_1,S_2)$ and
$\Var(\hat S_i)=\Var(S_i)/r_i$, proving the first identity.
Independent replicates have covariance $\Var(S_i)$ and variance
$\Var(\hat S_i)$, hence correlation $r_i$. Consistency follows by
continuous mapping. \qed

\begin{corollary}[Stability extends to unbalanced plans]\label{cor:unbal}
Fix $\varepsilon,\lambda_A,\lambda_B>0$, a reference $R$ and marginals
$\mu_A,\mu_B$ with full support on the retained sets, and let $\pi,\hat\pi$
minimize
\[
F_c(\pi):=\langle c,\pi\rangle+\varepsilon\,\KL_g(\pi\|R)
+\lambda_A\,\KL_g(\pi\mathbf1\|\mu_A)+\lambda_B\,\KL_g(\pi^\top\mathbf1\|\mu_B)
\]
over $\pi\ge0$, for costs $c$ and $\hat c$ with $\delta:=\|\hat c-c\|_\infty$,
where $\KL_g(p\|q):=\sum_i\big[p_i\log(p_i/q_i)-p_i+q_i\big]$ is the
generalized divergence of nonnegative vectors. Then
\[
\varepsilon\,\KL_g(\hat\pi\|\pi)\le\delta\,\|\hat\pi-\pi\|_1
\qquad\text{and}\qquad
\|\hat\pi-\pi\|_1\ \le\ \frac{2\delta}{3\varepsilon}\big(\|\hat\pi\|_1+2\|\pi\|_1\big),
\]
which reduces to the $2\delta/\varepsilon$ bound of Theorem~\ref{thm:stab} when
both plans have unit mass.
\end{corollary}

\paragraph{Proof of Corollary~\ref{cor:unbal}.}
At a zero entry on $\supp R$, the entropy derivative is $-\infty$, the
cost derivative is finite, and the marginal-penalty derivatives are finite
or $-\infty$. Thus the optimum is interior and $\nabla F_c(\pi)=0$.
Its objective gap is the sum of the plan and marginal Bregman divergences,
giving $F_c(\pi')-F_c(\pi)\ge\varepsilon\KL_g(\pi'\|\pi)$.
Optimality under $\hat c$ then gives
$\varepsilon\KL_g(\hat\pi\|\pi)\le F_c(\hat\pi)-F_c(\pi)
=\big[F_{\hat c}(\hat\pi)-F_{\hat c}(\pi)\big]
+\langle c-\hat c,\hat\pi-\pi\rangle\le\delta\|\hat\pi-\pi\|_1$.
For nonnegative $p,q$, the generalized Pinsker inequality is
\[
\KL_g(p\|q)\ \ge\ \frac{3\,\|p-q\|_1^2}{2\,\big(\|p\|_1+2\|q\|_1\big)} .
\]
Indeed, $u\log u-u+1\ge3(u-1)^2/(2(u+2))$ for $u\ge0$: the difference
and its derivative vanish at one, while its second derivative is
$1/u-27/(u+2)^3\ge0$ by AM--GM. For $u_i=p_i/q_i$, Cauchy--Schwarz gives
\[
\|p-q\|_1=\sum_iq_i|u_i-1|
=\sum_i\sqrt{q_i(u_i+2)}\;\sqrt{\frac{q_i(u_i-1)^2}{u_i+2}}
\le\sqrt{\|p\|_1+2\|q\|_1}\;\sqrt{\tfrac23\,\KL_g(p\|q)} .
\]
Zero coordinates follow by limits, with infinite divergence if $p_i>q_i=0$.
Combining the two bounds with $p=\hat\pi$, $q=\pi$ gives
$\varepsilon\cdot\frac{3\|\hat\pi-\pi\|_1^2}{2(\|\hat\pi\|_1+2\|\pi\|_1)}
\le\varepsilon\KL_g(\hat\pi\|\pi)\le\delta\|\hat\pi-\pi\|_1$.
Dividing when the distance is positive proves the claim; unit masses give
$2\delta/\varepsilon$. \qed

\subsection{Geometry Bounds for Overcomplete Dictionaries}\label{app:ceiling}

\begin{assumption}[Proportional-amplitude absorption model]\label{ass:abs}
Consider the pair of features in Appendix~\ref{app:canonical},
Example~2, with supports
$S\subseteq G$ satisfying $\Pr(S)>0$ and $\Pr(G\setminus S)>0$; amplitudes
$h_{\mathrm{gen}},h_{\mathrm{sp}}\ge0$ with $h_{\mathrm{sp}}=0$ off $S$,
$h_{\mathrm{gen}}=c\,h_{\mathrm{sp}}$ on $S$ for a constant $c>0$, and
$\E h_{\mathrm{sp}}>0$; orthogonal unit-norm decoders
$d_{\mathrm{gen}}\perp d_{\mathrm{sp}}$; training objective equal to
reconstruction error plus an $\ell_1$ penalty with coefficient
$\lambda>0$.
\end{assumption}

\begin{lemma}[Bounds from squared decoder cosines]\label{lem:ceiling}

Let $D\in\R^{n\times m}$ have unit-norm columns, with $m\ge2$, and let
$\sigma_1,\dots,\sigma_n$ be its singular values. The mean squared cosine
between distinct decoder columns satisfies the identity
\[
\overline{s_\dec^2}
:=\frac{1}{m(m-1)}\sum_{j\ne k}s_\dec(j,k)^2
=\frac{\sum_i\sigma_i^4-m}{m(m-1)}.
\]
For every $\tau>0$, Markov's inequality implies that the fraction of
unordered pairs of distinct columns with
$|s_\dec|\ge\tau$ is at most $\overline{s_\dec^2}/\tau^2$; equivalently, at
most $(m-1)\overline{s_\dec^2}/\tau^2$ partners per feature on average.
Both upper bounds are computable from $D$ alone.

\end{lemma}

\paragraph{Proof of Lemma~\ref{lem:ceiling}.}

The identity follows from
$\sum_{j,k}\langle d_j,d_k\rangle^2=\|D^\top D\|_F^2
=\operatorname{tr}\big((D^\top D)^2\big)=\sum_i\sigma_i^4$, and the diagonal
contribution $m$. Markov's inequality for $s_\dec^2$ under uniform sampling
of distinct pairs gives the fraction bound; multiplying by $m-1$ gives
the average partner count. \qed

\begin{remark}[Welch lower bound]\label{rem:welch}

$\overline{s_\dec^2}\ge(m-n)/(n(m-1))\sim1/n$, with equality in the inequality
iff $D$ is a tight frame, $DD^\top=(m/n)I$
\citep{welch1974lower,benedetto2003finite}.
The equivalence holds as $m/n\to\infty$.
The proof is trivial: the eigenvalues of $DD^\top$ satisfy $\sum_i\lambda_i=m$;
Cauchy--Schwarz gives $\sum_i\lambda_i^2\ge m^2/n$, with equality iff
all $\lambda_i=m/n$. Substitute into Lemma~\ref{lem:ceiling}.
\end{remark}

\begin{corollary}[Upper bounds under approximate tightness]\label{cor:tight}
If $\opn{(n/m)DD^\top-I}\le\etafr$, then
\[
\overline{s_\dec^2}\le
\frac{m(1+\etafr)-n}{n(m-1)},\qquad
\Pr[|s_\dec|\ge\tau]\le
\frac{m(1+\etafr)-n}{n(m-1)\tau^2}.
\]
Here, the probability is under uniform sampling of distinct decoder
pairs, as in Lemma~\ref{lem:ceiling}.
\end{corollary}
\begin{proof}
Every eigenvalue of $DD^\top$ is at most $(m/n)(1+\etafr)$.
Consequently,
$\sum_i\lambda_i^2\le(\max_i\lambda_i)\sum_i\lambda_i
\le(m^2/n)(1+\etafr)$.
Substitution in Lemma~\ref{lem:ceiling} and Markov's inequality prove
the bounds.
\end{proof}

\begin{corollary}[Energy-weighted upper bound]\label{cor:ceiling-energy}

Let $\bar w$ be the normalized energy weights and
$\overline W:=\mathrm{diag}(\bar w)$. Under the product measure
$\bar w\otimes\bar w$, the mass of distinct decoder pairs with absolute
cosine at least $\tau>0$ satisfies
\[
(\bar w\otimes\bar w)\big(\{(j,k):j\ne k,\ |s_\dec(j,k)|\ge\tau\}\big)
\ \le\ \frac{\operatorname{tr}\big((D\overline WD^\top)^2\big)-\sum_j\bar w_j^2}{\tau^2}.
\]
The right-hand side is computable from $D$ and the weights alone.

\end{corollary}

\paragraph{Proof of Corollary~\ref{cor:ceiling-energy}.}
Expand
$\operatorname{tr}((D\overline WD^\top)^2)
=\sum_{j,k}\bar w_j\bar w_k\langle d_j,d_k\rangle^2$,
subtract the diagonal $\sum_j\bar w_j^2$, and apply Markov's inequality
under $\bar w\otimes\bar w$. \qed

\begin{proposition}[A lower sparsity penalty under absorption]\label{prop:forced}
Under Assumption~\ref{ass:abs}, the absorbed
configuration of Appendix~\ref{app:absorption-model} reconstructs the
same signal as the disentangled one while lowering the objective by
\[
\Delta L=\lambda\big[(1+c)-\sqrt{1+c^2}\big]\,\E[h_{\mathrm{sp}}\1_S]\ >\ 0
\quad\text{for every }\lambda>0.
\]
The decoder cosine is
$s_\dec(d_{\mathrm{sp}},d_{\mathrm{abs}})=1/\sqrt{1+c^2}$.
The absorbed energy fraction $\varphi=c^2\E[h_{\mathrm{sp}}^2]/w_g$
is fixed by co-occurrence statistics and is $\Theta(1)$ as $m,n$ vary
with those statistics held fixed; here $w_g=\E h_{\mathrm{gen}}^2$.
\end{proposition}

\paragraph{Interpretation.}
The objective-based explanation of absorption \citep{dorrell2026optimality} is explicit here for the two specified configurations. Identifying a transport fraction with absorbed energy requires the isolated model of Proposition~\ref{prop:plan-abs} and $\varepsilon\to0$. At fixed $\varepsilon>0$, the CLTs quantify sampling uncertainty for the regularized transport fractions.

\paragraph{Proof of Proposition~\ref{prop:forced}.}

Off $S$ the configurations agree. On $S$, write $b=h_{\rm sp}$:
both reconstruct $cb\,d_{\rm gen}+b\,d_{\rm sp}$, with respective
$\ell_1$ costs $(1+c)b$ and $\sqrt{1+c^2}\,b$.
Since $(1+c)^2-(1+c^2)=2c>0$ and $\E h_{\rm sp}>0$, averaging gives
the stated positive objective difference. Orthogonality gives
$\langle d_{\rm sp},d_{\rm abs}\rangle=1/\sqrt{1+c^2}$.
Finally, $\E[h_{\rm gen}^2\1_S]=c^2\E h_{\rm sp}^2$ and splitting
$\E h_{\rm gen}^2$ over $S$ and $G\setminus S$ gives $\varphi$. \qed

\begin{remark}[\nrev{Pair counts, product measures, and couplings}]\label{rem:count-energy}
For aligned pairs $Q$,
Lemma~\ref{lem:ceiling} bounds their frequency under uniform sampling;
Corollary~\ref{cor:ceiling-energy} bounds their mass under independent
energy-weighted sampling. The corresponding maximum over couplings is
$\max_{\pi\in\Pi(\mu_A,\mu_B)}\pi(Q)$, which allows dependence subject
to the marginal constraints.
\end{remark}

\section{Statistical Inference (Section~\ref{sec:fss})}

\subsection{Derivatives and Central Limit Theorems}\label{app:clt-proof}

\paragraph{Moments and sampling.}\label{app:moment-conventions}
For $Y=(U,V,\operatorname{vec}W)$ and
$\vartheta=\E Y=(u,v,\operatorname{vec}w)$, define
\[
\begin{gathered}
E_A:=\sum_j u_j,\quad E_B:=\sum_k v_k,\quad
a_j:=u_j/E_A,\quad b_k:=v_k/E_B,\quad
s_{jk}:=w_{jk}/\sqrt{u_jv_k},\quad c_{jk}:=2(1-s_{jk}).
\end{gathered}
\]
We further condition on the trained SAEs and independent selection data.
Finite fourth activation moments averaged over documents suffice for Assumption~\ref{ass:document-clt}.
We flag samples with zero retained energy as invalid.
\begin{lemma}[Smoothness of the entropic plan]\label{lem:plan-smoothness}
For fixed $\varepsilon>0$, positive probability marginals, and finite costs, the
entropic plan is positive and smooth in the costs and marginals.
Consequently, its entries and transport cost are smooth functions of
the moment parameters wherever retained energies are positive, as are
fractions with $a(I)>0$.
\end{lemma}
\begin{proof}
Mixing a plan with $a\otimes b$ gives entropy derivative $-\infty$ at
any zero cell, so an optimum is positive. Write $p=\operatorname{vec}P$
and let $A_0$ encode all row sums and all but one column sum.
On each coupling polytope, the KL objective differs by a constant from
$\langle c,p\rangle+\varepsilon\sum_\ell p_\ell(\log p_\ell-1)$.
Its stationarity and feasibility equations have Jacobian
\[
\begin{pmatrix}
\varepsilon\operatorname{diag}(1/p)&-A_0^\top\\
A_0&0
\end{pmatrix}.
\]
A null vector satisfies $A_0dp=0$ and
$\varepsilon\sum_\ell(dp_\ell)^2/p_\ell=0$, hence $dp=0$.
Full row rank of $A_0$ gives $d\lambda=0$.
The implicit function theorem and the moment formulas above prove
smoothness, following the sensitivity argument of
\citet{klatt2020regularized}.
\end{proof}

\paragraph{Derivative of a transport fraction.}
For $\theta=P(I\times J)/a(I)$, differentiation gives
\[
d\theta=\langle g,dP\rangle,\qquad
g_{jk}=\frac{\1_I(j)(\1_J(k)-\theta)}{a(I)}.
\]
Project $g$ onto additive functions in $L^2(P)$, writing
$g^\perp_{jk}=g_{jk}-\alpha_j-\beta_k$. The projection equations are
\begin{equation}
\begin{gathered}
H_P\begin{pmatrix}\alpha\\\beta\end{pmatrix}
=\begin{pmatrix}
(\sum_kP_{jk}g_{jk})_j\\
(\sum_jP_{jk}g_{jk})_k
\end{pmatrix},\quad
H_P=\begin{pmatrix}\operatorname{diag}(a)&P\\
P^\top&\operatorname{diag}(b)\end{pmatrix}.
\end{gathered}
\label{eq:clt-projection}
\end{equation}
Thus $\sum_kP_{jk}g^\perp_{jk}=\sum_jP_{jk}g^\perp_{jk}=0$.
The fitted sum is unique: since $P_{jk}>0$,
$z^\top H_Pz=\sum_{jk}P_{jk}(z_j+z_{K_A+k})^2$
vanishes only for $z=t(\mathbf1,-\mathbf1)$.

Differentiating the Gibbs representation gives
$\varepsilon\,dP_{jk}/P_{jk}=df_j+dg_k-dc_{jk}$.
Orthogonality to additive functions yields
\begin{equation}
d\theta=\alpha^\top da+\beta^\top db
-\varepsilon^{-1}\sum_{jk}P_{jk}g^\perp_{jk}\,dc_{jk}.
\label{eq:clt-differential}
\end{equation}
Substitute the moment derivatives
\[
\begin{aligned}
da_j&=\frac{du_j-a_j\sum_\ell du_\ell}{E_A},\quad
db_k=\frac{dv_k-b_k\sum_\ell dv_\ell}{E_B},\quad
dc_{jk}&=-2\left\{
\frac{dw_{jk}}{\sqrt{u_jv_k}}
-\frac{s_{jk}}2\left(\frac{du_j}{u_j}+\frac{dv_k}{v_k}\right)
\right\}.
\end{aligned}
\]
With $\bar\alpha=\sum_j a_j\alpha_j$ and
$\bar\beta=\sum_k b_k\beta_k$, the influence function is
\begin{align}
\psi(X)
={}&\sum_j\frac{\alpha_j-\bar\alpha}{E_A}\,U_j(X)
 +\sum_k\frac{\beta_k-\bar\beta}{E_B}\,V_k(X)+\frac2\varepsilon\sum_{jk}P_{jk}g^\perp_{jk}
\Biggl\{\frac{W_{jk}(X)}{\sqrt{u_jv_k}}-\frac{s_{jk}}2\left(\frac{U_j(X)}{u_j}
                    +\frac{V_k(X)}{v_k}\right)\Biggr\}.
\label{eq:clt-influence}
\end{align}
It has mean zero and is invariant under
$(\alpha,\beta)\mapsto(\alpha+t\mathbf1,\beta-t\mathbf1)$.

\begin{proof}[Proof of Theorem~\ref{thm:clt}]
Assumption~\ref{ass:document-clt} gives
$\sqrt N(\hat\vartheta-\vartheta)\Rightarrow\mathcal N(0,\Omega)$,
where $\Omega=\Cov(Y)$ is finite. Lemma~\ref{lem:plan-smoothness}
and the delta method give
\[
\mathcal T_\varepsilon(\hat\vartheta)
-\mathcal T_\varepsilon(\vartheta)
=J_\varepsilon(\hat\vartheta-\vartheta)+o_p(N^{-1/2}),
\]
proving \eqref{eq:clt-linear}--\eqref{eq:clt-limit} with
$\Sigma=J_\varepsilon\Omega J_\varepsilon^\top$.
The sample covariance $\hat\Omega$ is consistent by finite second
moments. Derivative continuity gives
\[
\begin{aligned}
\hat\Sigma
=\frac1{N-1}\sum_i(\hat\psi_i-\bar{\hat\psi})
(\hat\psi_i-\bar{\hat\psi})^\top=\hat J_\varepsilon\hat\Omega\hat J_\varepsilon^\top\to_p\Sigma.
\end{aligned}
\]
For $\Sigma_{rr}>0$, Slutsky's theorem yields the studentized normal
limit. A union bound over fixed $F$ intervals, each with limiting
noncoverage $\alpha/F$, gives simultaneous coverage at least
$1-\alpha$. Intersecting with $[0,1]$ preserves coverage.
\end{proof}

\paragraph{Selection and the two-topic protocol.}\label{app:clt-selection}
Allocate $\alpha/2$ to simultaneous inference on each topic.
After inspecting topic A, select a subset $\mathcal R_A$ of the fixed
$F$ fractions for evaluation on independent topic B data.
Require positive limiting variances for all A coordinates and all
potentially reported B coordinates; report exact identities directly.
For every nonempty subset, Bonferroni correction by $|\mathcal R_A|$
bounds asymptotic B noncoverage by $\alpha/2$.
There are finitely many subsets, so this bound is uniform over the A
decision. Independence permits conditioning on that decision.
Combining with the A simultaneous event gives joint asymptotic coverage
at least $1-\alpha$ as both sample sizes grow.

\paragraph{Cost thresholds and total variation.}\label{app:clt-nonsmooth}
For a fixed threshold $\delta$, define
\[
J_j(\delta)=\{k:c_{jk}\le\delta\},\qquad
\hat J_j(\delta)=\{k:\hat c_{jk}\le\delta\}.
\]
On a fixed finite dictionary, assume
\[
\gamma_\delta:=\min_{j,k}|c_{jk}-\delta|>0,
\qquad \|\hat c-c\|_\infty\xrightarrow{p}0.
\]
Then
\[
\begin{aligned}
\Pr\{\hat J_j(\delta)=J_j(\delta)\ \forall j\}
&\ge \Pr\{\|\hat c-c\|_\infty<\gamma_\delta\}\\
&\longrightarrow 1.
\end{aligned}
\]
Thus Theorem~\ref{thm:clt} applies to the fractions
\[
\theta_j=\frac{\sum_{k\in J_j(\delta)}P_{jk}}{a_j},
\qquad
\hat\theta_j=
\frac{\sum_{k\in\hat J_j(\delta)}\hat P_{jk}}{\hat a_j}.
\]
For independently selected sets, the theorem applies conditionally
on selection. Lemma~\ref{lem:sandwich} gives deterministic bounds
when separate error bounds are available.

For two smooth plans based on the same moments, let
$D_{jk}=P^{(1)}_{jk}-P^{(2)}_{jk}$ and let $Z$ be the Gaussian limit of
$\sqrt N(\hat D-D)$. The directional delta method gives
\[
\sqrt N(\widehat{\Dis}-\Dis)\Rightarrow
\frac12\left\{
\sum_{D_{jk}\ne0}\operatorname{sign}(D_{jk})Z_{jk}
+\sum_{D_{jk}=0}|Z_{jk}|
\right\}.
\]
Without zero entries, this limit gives a normal interval when its
variance is positive. At zero entries, Corollary~\ref{cor:bootstrap-radius}
uses the norm of the vector estimation error to obtain a conservative
interval.

\subsubsection{Transport Costs and Joint Inference}\label{app:cost-clt-proof}

\paragraph{Derivative of the transport cost.}
For $P=P^{st}_\varepsilon$, solve
\begin{equation}
\begin{pmatrix}\operatorname{diag}(a)&P\\P^\top&\operatorname{diag}(b)\end{pmatrix}
\binom{\lambda_A}{\lambda_B}
=\binom{(P\odot c)\mathbf1}{(P\odot c)^\top\mathbf1}.
\label{eq:cost-adjoint}
\end{equation}
Here $\odot$ denotes elementwise multiplication; fix one coefficient
to identify the solution. Define
\[
G_{jk}=P_{jk}\left(1+\frac{\lambda_{A,j}+\lambda_{B,k}-c_{jk}}{\varepsilon}\right).
\]
Then
\begin{equation}
dD_\varepsilon
=\lambda_A^\top da+\lambda_B^\top db+\langle G,dc\rangle.
\label{eq:cost-differential}
\end{equation}
Indeed, $r_{jk}=c_{jk}-\lambda_{A,j}-\lambda_{B,k}$ satisfies
$\sum_kP_{jk}r_{jk}=\sum_jP_{jk}r_{jk}=0$ by
\eqref{eq:cost-adjoint}. The Gibbs derivative therefore gives
\[
\begin{aligned}
d\langle c,P\rangle
=\langle P,dc\rangle+\lambda_A^\top da+\lambda_B^\top db
  +\sum_{jk}r_{jk}\,dP_{jk} =\lambda_A^\top da+\lambda_B^\top db
  +\sum_{jk}P_{jk}(1-r_{jk}/\varepsilon)\,dc_{jk}.
\end{aligned}
\]
Since $\sum_jda_j=\sum_kdb_k=0$, the result is invariant to the
coefficient normalization.

For activation costs, define the moment coefficients
\[
\begin{aligned}
A_j=\frac{\lambda_{A,j}-a^\top\lambda_A}{E_A}
       +\frac1{u_j}\sum_kG_{jk}s_{jk},\quad
B_k=\frac{\lambda_{B,k}-b^\top\lambda_B}{E_B}
       +\frac1{v_k}\sum_jG_{jk}s_{jk},\quad
K_{jk}=-\frac{2G_{jk}}{\sqrt{u_jv_k}}.
\end{aligned}
\]
Substituting the moment fluctuations gives the mean-zero influence
function
\begin{equation}
\begin{aligned}
\chi(X)=\sum_j A_j(U_j(X)-u_j)
       +\sum_k B_k(V_k(X)-v_k)+\sum_{jk}K_{jk}(W_{jk}(X)-w_{jk}).
\end{aligned}
\label{eq:cost-influence}
\end{equation}
For fixed decoder costs, $dc=0$, so only the two marginal derivatives
contribute.

\begin{proof}[Proof of Theorem~\ref{thm:cost-clt}]
Lemma~\ref{lem:plan-smoothness} makes
$\mathcal F:\vartheta^{\rm all}\mapsto\mathbf D_\varepsilon$ smooth.
For $J=D\mathcal F(\vartheta^{\rm all})$ and
$V=\Cov(Y^{\rm all})$, the mean CLT and delta method give
\eqref{eq:cost-linear}--\eqref{eq:cost-clt}, with
$\boldsymbol\chi=J(Y^{\rm all}-\vartheta^{\rm all})$ and
$\Omega=JVJ^\top$. Shared energy coordinates occur once in
$Y^{\rm all}$ and affect every relevant pair, retaining their covariances.
Finite second moments and derivative continuity imply
$\hat\Omega=\hat J\hat V\hat J^\top\to_p\Omega$.
Slutsky's theorem gives the interval whenever $v^\top\Omega v>0$,
including singular $\Omega$.
Appending the fraction coordinates to $\mathcal F$ gives joint
inference for fractions and costs, with covariance estimated by
stacking $\psi$ and $\boldsymbol\chi$ for each document.
\end{proof}

\begin{proof}[Proof of Corollary~\ref{cor:context-cost}]
Conditional on the SAEs, the topic samples give independent Gaussian
limits for $\sqrt{N_g}(\hat M_g-M_g)$ and
$\sqrt{N_h}(\hat M_h-M_h)$.
The sample-size ratio condition and positive variances give a standard
normal limit for their centered difference divided by
$\sqrt{v^\top\Omega_gv/N_g+v^\top\Omega_hv/N_h}$.
Covariance consistency and Slutsky's theorem justify the estimated
denominator.
\end{proof}

\subsection{Joint Document Bootstrap}\label{ssec:bootstrap}\label{app:bootstrap-proof}

Let $Y_i=Y^{\rm all}(X_i)$ contain each required energy and cross-moment
once, with mean $\vartheta$. Let $\mathcal H(\vartheta)\in\R^K$ collect
a fixed finite set of fractions, costs, plan entries, or fixed linear
combinations, possibly the entire plan. Define
\[
\begin{aligned}
\hat H&=\mathcal H(\hat\vartheta),& J_H&=D\mathcal H(\vartheta),\\
\Psi_i&=J_H(Y_i-\vartheta),& \Sigma_H&=\E\Psi_i\Psi_i^\top,
\end{aligned}
\]
and $\hat\Psi_i=D\mathcal H(\hat\vartheta)(Y_i-\hat\vartheta)$,
$\hat\Sigma_H=(N-1)^{-1}\sum_i\hat\Psi_i\hat\Psi_i^\top$.

Resample $N$ document indices uniformly with replacement, using the
same indices for all SAE pairs. Recompute moments, costs, marginals,
and plans to obtain $\hat H^*=\mathcal H(\hat\vartheta^*)$ and
$\hat\Sigma_H^*$. For zero-energy resamples, set
$\hat H^*=\hat H$, $\hat\Sigma_H^*=\hat\Sigma_H$, and record their frequency.

\begin{theorem}[Joint document and multiplier bootstrap]\label{thm:bootstrap}
Under Assumption~\ref{ass:document-clt} for $Y^{\rm all}$,
\[
\sqrt N(\hat H-\mathcal H(\vartheta))\Rightarrow
Z_H\sim\mathcal N_K(0,\Sigma_H),\qquad \hat\Sigma_H\to_p\Sigma_H.
\]
For bounded-Lipschitz distance $d_{\rm BL}$ and conditional resampling
law $\mathcal L^*$,
\begin{equation}
d_{\rm BL}\!\left(\mathcal L^*\{\sqrt N(\hat H^*-\hat H)\},
                   \mathcal N_K(0,\Sigma_H)\right)\to_p0.
\label{eq:bootstrap-law}
\end{equation}
Also, $\hat\Sigma_H^*\to_{P^*}\Sigma_H$ in probability: for every
$\eta>0$, $\Pr^*(\|\hat\Sigma_H^*-\Sigma_H\|>\eta)\to_p0$.
For independent $\xi_i\sim\mathcal N(0,1)$, define
$Z_N^{\rm mult}=\frac1{\sqrt{N-1}}\sum_i\xi_i\hat\Psi_i,$
using the same multiplier for every coordinate from a given document.
Its conditional law is $\mathcal N_K(0,\hat\Sigma_H)$ and
converges to the same limit in $d_{\rm BL}$.
Singular covariance is allowed; fixed decoder costs have zero cost
derivatives.
\end{theorem}
\begin{proof}
By Lemma~\ref{lem:plan-smoothness}, $\mathcal H$ is smooth on an open
positive-energy neighborhood of $\vartheta$, including when the moment
matrix is singular. The mean CLT and delta method give
\[
\hat H-\mathcal H(\vartheta)
=J_H(\hat\vartheta-\vartheta)+o_p(N^{-1/2}),\qquad
\Sigma_H=J_HVJ_H^\top,
\]
where $V=\Cov(Y_i)$. Finite second moments give sample covariance
consistency, $\hat V\to_pV$, and derivative continuity yields
$\hat\Sigma_H=D\mathcal H(\hat\vartheta)\hat V
D\mathcal H(\hat\vartheta)^\top\to_p\Sigma_H$.

Conditionally, $R_N^*=\sqrt N(\hat\vartheta^*-\hat\vartheta)$ is a
normalized sum of centered empirical draws with covariance
\[
V_N^{\rm emp}=N^{-1}\sum_i(Y_i-\hat\vartheta)
(Y_i-\hat\vartheta)^\top\to_pV.
\]
For every $\eta>0$, its Lindeberg quantity is $L_N(\eta)=N^{-1}\sum_i\|Y_i-\hat\vartheta\|^2
\1\{\|Y_i-\hat\vartheta\|>\eta\sqrt N\}.$
On $\|\hat\vartheta\|\le\eta\sqrt N/2$, an event of probability
tending to one, it is bounded by
$4N^{-1}\sum_i\|Y_i\|^2\1\{\|Y_i\|>\eta\sqrt N/2\}$.
Integrability makes this bound tend to zero in probability.
The conditional Lindeberg--Feller theorem and Cram\'er--Wold device
therefore give the Gaussian limit, including singular $V$.
Formally, every subsequence has a further subsequence with almost-sure
covariance and Lindeberg convergence for all rational $\eta>0$.

Since $\E^*\|R_N^*\|^2=\operatorname{tr}V_N^{\rm emp}=O_p(1)$,
the resampled mean remains in the smooth neighborhood with conditional
probability tending to one. The mean-value expansion gives
\[
\sqrt N(\hat H^*-\hat H)=J_HR_N^*+o_{P^*}(1)
\]
in probability, proving \eqref{eq:bootstrap-law}.
For each retained energy coordinate $A_\ell\ge0$, put
$p_\ell=\Pr(A_\ell>0)>0$ and
$\hat p_\ell=N^{-1}\sum_i\1\{A_{\ell,i}>0\}$.
The zero-energy probabilities are $(1-p_\ell)^N$ and, conditionally,
$(1-\hat p_\ell)^N\le e^{-N\hat p_\ell}\to_p0$.
A finite union justifies the stated extensions.

To prove resample covariance consistency, truncate each integrable
product $Y_jY_k$ at magnitude $M$. Its resampled average has conditional
variance $O(M^2/N)$, and its expected absolute tail is the empirical
tail. The law of large numbers, then $M\to\infty$, gives conditional
consistency of raw second moments. Together with mean consistency and
derivative continuity, this proves
$\hat\Sigma_H^*\to_{P^*}\Sigma_H$ in probability.
Finally, the multiplier vector is conditionally Gaussian with covariance
$\hat\Sigma_H$. Continuity of the nonnegative matrix square root gives
convergence even when $\Sigma_H$ is singular.
\end{proof}

\begin{corollary}[Scalar and simultaneous intervals]\label{cor:bootstrap-ci}
Assume Theorem~\ref{thm:bootstrap}. For fixed $v$ with
$\sigma_v^2=v^\top\Sigma_Hv>0$, let
$\hat\sigma_v^2=v^\top\hat\Sigma_Hv$ and let $q_p^*$ be the conditional
$p$-quantile of $T_v^*=\frac{\sqrt N\,v^\top(\hat H^*-\hat H)}
{\sqrt{v^\top\hat\Sigma_H^*v}}.$
Set $T_v^*=0$ on zero-energy resamples or when its denominator vanishes.
Then $\left[v^\top\hat H-q_{1-\alpha/2}^*\frac{\hat\sigma_v}{\sqrt N},\quad
v^\top\hat H-q_{\alpha/2}^*\frac{\hat\sigma_v}{\sqrt N}\right]$
has limiting coverage $1-\alpha$. Using $\hat\sigma_v$ in the bootstrap
denominator gives the same first-order validity.

For simultaneous inference, assume $\Sigma_{H,rr}>0$ for every included
coordinate and set $\hat\sigma_r^2=\hat\Sigma_{H,rr}$.
Let $q^*$ be the conditional $(1-\alpha)$-quantile of either
\[
\max_r\frac{|\sqrt N(\hat H_r^*-\hat H_r)|}{\hat\sigma_r}
\quad\text{or}\quad
\max_r\frac{|Z_{N,r}^{\rm mult}|}{\hat\sigma_r}.
\]
The intervals $\hat H_r\pm q^*\hat\sigma_r/\sqrt N$ then have simultaneous
limiting coverage $1-\alpha$; fraction intervals may be clipped to $[0,1]$.
For original samples with zero energies or required empirical variances,
use a finite measurable convention and flag the report as invalid.
These events have vanishing probability.
Empirical quantiles retain these limits for $B_N\to\infty$ conditionally
i.i.d. draws, at any rate relative to $N$.
\end{corollary}
\begin{proof}
Conditional Slutsky gives $T_v^*\Rightarrow\mathcal N(0,1)$ and
convergence of its quantiles. The original standardized error has the
same limit, proving scalar coverage. Scaling by $\hat\sigma_v$ is
equivalent asymptotically. A standardized scalar Gaussian multiplier
is normal, giving the usual Wald interval.

For simultaneous intervals, let
$M=\max_r|Z_{H,r}|/\sqrt{\Sigma_{H,rr}}$.
Positive marginal variances exclude atoms: $\{M=t\}$ lies in finitely
many scalar Gaussian level sets. Every nonempty open norm annulus in
the Gaussian support has positive probability, so its CDF is strictly
increasing on $(0,\infty)$. Conditional continuous mapping and Slutsky
give quantile convergence and simultaneous coverage.
Clipping to a known parameter range preserves coverage.
The distribution-free empirical-CDF bound gives quantile consistency
whenever $B_N\to\infty$; fixed $B$ retains Monte Carlo error.
\end{proof}

\begin{corollary}[Confidence bounds for plans and their differences]\label{cor:bootstrap-radius}
Under Theorem~\ref{thm:bootstrap}, take $\mathcal H=\operatorname{vec}P$
with nonzero limiting covariance. Let $q_P^*$ be the conditional
$(1-\alpha)$-quantile of $\|\sqrt N(\hat P^*-\hat P)\|_1$ or its
multiplier counterpart. Then
\begin{equation}
\Pr\!\left\{\|\hat P-P\|_1\le
 \min\{2,q_P^*/\sqrt N\}\right\}\longrightarrow1-\alpha.
\label{eq:bootstrap-plan-radius}
\end{equation}
For two smooth plans on the same fixed cells, put
$\Delta=\operatorname{vec}(P^{(1)}-P^{(2)})$.
If its limiting covariance is nonzero, let $q_\Delta^*$ be the
corresponding quantile of
$\|\sqrt N(\hat\Delta^*-\hat\Delta)\|_1$ or its multiplier counterpart.
Then
\[
\left[\widehat{\Dis}-\frac{q_\Delta^*}{2\sqrt N},
      \widehat{\Dis}+\frac{q_\Delta^*}{2\sqrt N}\right]\cap[0,1]
\]
has $\liminf$ coverage at least $1-\alpha$ for
$\Dis=\|\Delta\|_1/2$, including cells with $\Delta_{jk}=0$.
\end{corollary}
\begin{proof}
The $\ell_1$ norm of a Gaussian vector with nonzero covariance has a
continuous, strictly increasing CDF on $(0,\infty)$: norm spheres have
zero measure on its support, and open annuli have positive measure.
Continuous mapping and quantile convergence give
\eqref{eq:bootstrap-plan-radius}. Capping at two preserves coverage.
For differences, use
\[
\left|\|\hat\Delta\|_1/2-\|\Delta\|_1/2\right|
\le\|\hat\Delta-\Delta\|_1/2.
\]
This bound applies also at zero cells, where the norm itself is
nondifferentiable. Calibration uses the centered vector error before
taking its norm.
\end{proof}

\paragraph{Numerical accuracy.}\label{app:bootstrap-numerics}
The same limits hold for computed vectors satisfying
$\sqrt N\|\widetilde H-\hat H\|\to_p0$ and, for every $\eta>0$,
\[
\Pr^*\{\sqrt N\|\widetilde H^*-\hat H^*\|>\eta\}\to_p0.
\]
Studentization also requires
$\widetilde\Sigma_H\to_p\Sigma_H$ and
$\widetilde\Sigma_H^*\to_{P^*}\Sigma_H$ in probability.
For multipliers, original-estimate accuracy and centered computed
influence vectors satisfying
\[
\frac1{N-1}\sum_i\|\widetilde\Psi_i-\hat\Psi_i\|^2\to_p0
\]
suffice. Under the same multipliers, this sum is the conditional mean
squared error of the multiplier vector. Markov's inequality makes it
negligible; Cauchy--Schwarz and $O_p(1)$ empirical influence second
moments give covariance consistency.

\begin{proposition}[Rare documents and finite-sample coverage]\label{prop:bootstrap-rare}
Take one position per document and activations
$h_A=h_{B,1}=1$, $h_{B,2}=1+MZ$, where
$Z\sim\operatorname{Bernoulli}(p)$, $0<p<1$, and $M>0$ are fixed.
All moments are finite and population energies positive. The transport
fraction to the first target is
\[
\theta=\frac1{2+p(2M+M^2)}<\frac12.
\]
With probability $(1-p)^N$, $\hat\theta=1/2$ and both bootstrap and
multiplier distributions are degenerate. An unflagged zero-width
plug-in interval then misses $\theta$, giving coverage at most
$1-(1-p)^N$.
\end{proposition}
\begin{proof}
One source row forces $P=(b_1,b_2)$ for every $\varepsilon>0$.
The target energies $v_1=1$, $v_2=1+p(2M+M^2)$ give the stated fraction.
If all observed $Z$ are zero, all moment vectors coincide:
$\hat\theta=1/2$, every resampled estimate equals $1/2$, and empirical
influence variance is zero. This event has probability $(1-p)^N$.
The population influence variance is positive because
$d\theta/dv_2=-1/(1+v_2)^2\ne0$ and the second target energy varies
across documents.
\end{proof}

\subsection{A Finite-Sample Plan Bound}\label{app:fss}

For independently selected latent sets, define
\[
g=(h^{(1)},h^{(2)}),\qquad
G(X)=L_X^{-1}\sum_t g(X,t)g(X,t)^\top,
\]
and $M=\E G(X)$, $\Mhat=N^{-1}\sum_iG(X_i)$.
The diagonal entries are energies; the cross block contains
co-activation moments. We first bound their estimation errors,
then propagate those bounds through the transport problem.

\begin{lemma}[Simultaneous moment bounds]\label{lem:fss-conc}
Let $X_1,\ldots,X_N$ be i.i.d. documents with positive retained lengths.
Suppose finite deterministic $V_d,V_x$ bound, respectively,
\[
\sum_{\ell\in\mathsf A\cup\mathsf B}\Var(G_{\ell\ell}(X)),\qquad
\sum_{j\in\mathsf A,k\in\mathsf B}\Var(G_{jk}(X)).
\]
For $\alpha\in(0,1)$, set $\eta_d=\sqrt{\frac{2V_d}{N\alpha}},
\eta_x=\sqrt{\frac{2V_x}{N\alpha}}.$
With probability at least $1-\alpha$,
\begin{equation}
\begin{gathered}
\max_\ell|\Mhat_{\ell\ell}-M_{\ell\ell}|\le\eta_d,\qquad
\max_{j,k}|\Mhat_{jk}-M_{jk}|\le\eta_x.
\end{gathered}
\label{eq:fss-moment-event}
\end{equation}
A zero variance sum makes the corresponding empirical entries exact
almost surely.
\end{lemma}
\begin{proof}
Each empirical average has variance $\Var(G_{jk}(X))/N$.
Chebyshev's inequality and a union bound give diagonal failure
probability at most $V_d/(N\eta_d^2)=\alpha/2$ when $V_d>0$,
and the same bound for the cross block when $V_x>0$.
A zero variance sum makes every entry in that block constant almost
surely. A final union bound proves the claim.
\end{proof}

For an unknown population, tail assumptions can supply valid bounds.
Any radii satisfying \eqref{eq:fss-moment-event} may be used below.

\begin{lemma}[Sensitivity of costs and marginals]\label{lem:fss-sens}
Let $\emin>0$ bound all retained population energies from below.
On \eqref{eq:fss-moment-event}, if $\eta_d\le\emin/2$, then
\[
\begin{gathered}
\|\hat a-a\|_1+\|\hat b-b\|_1\le8\eta_d/\emin,\quad
\max_{j,k}|\hat s_{jk}-s_{jk}|\le2(\eta_x+\eta_d)/\emin.
\end{gathered}
\]
Each marginal satisfies the bound $4\eta_d/\emin$;
activation costs satisfy
$\|\hat c-c\|_\infty\le4(\eta_x+\eta_d)/\emin$.
\end{lemma}
\begin{proof}
For one SAE, put $w_\ell=M_{\ell\ell}$,
$\hat w_\ell=\Mhat_{\ell\ell}$, $W=\sum w_\ell$, and
$\hat W=\sum\hat w_\ell$. With $K$ retained latents,
\[
\left\|\frac{\hat w}{\hat W}-\frac{w}{W}\right\|_1
\le\frac{2\sum_\ell|\hat w_\ell-w_\ell|}{\hat W}
\le\frac{2K\eta_d}{K\emin/2}=\frac{4\eta_d}{\emin}.
\]
For a cosine, changing the numerator contributes at most
$2\eta_x/\emin$. Positive semidefiniteness gives
$|M_{jk}|\le\sqrt{M_{jj}M_{kk}}$.
With $t=2\eta_d/\emin\le1$, each ratio
$M_{\ell\ell}/\Mhat_{\ell\ell}$ lies in $[1-t,1+t]$,
as does their geometric mean. The denominator change therefore
contributes at most $t$. Doubling the cosine bound gives the cost bound.
\end{proof}

\begin{lemma}[A nearby coupling with prescribed marginals]\label{lem:repair}
For a coupling $Q$ of $(\hat a,\hat b)$ and probability vectors $(a,b)$,
there is a coupling $R$ of $(a,b)$ with
\[
\|R-Q\|_1\le\|\hat a-a\|_1+\|\hat b-b\|_1.
\]
\end{lemma}
\begin{proof}
Scale down rows of $Q$ to the componentwise minima of $\hat a$ and $a$,
removing mass $t_A=\|\hat a-a\|_1/2$.
Trim columns to at most $b$, removing at most
$t_B=\|\hat b-b\|_1/2$ more.
The remaining row and column deficits have equal total mass
$\gamma\le t_A+t_B$. Add their outer product divided by $\gamma$
(nothing if $\gamma=0$).
The resulting coupling has marginals $(a,b)$ and differs from $Q$
by at most $2\gamma\le2t_A+2t_B$ in $\ell_1$.
\end{proof}

\begin{lemma}[Stability under marginal changes]\label{lem:fss-marg}
For fixed $\varepsilon>0$, cost $0\le c\le\bar c$, and positive
probability marginals, let $P,Q$ be the entropic optima for $(a,b)$ and
$(\hat a,\hat b)$. Set
$d=\min\{\|\hat a-a\|_1+\|\hat b-b\|_1,2\}$ and $D=K_AK_B$.
Then
\[
\|Q-P\|_1\le d+
2\sqrt{\frac{d\{\bar c+\varepsilon\log(eD/d)\}}{\varepsilon}},
\]
with right-hand side zero when $d=0$.
\end{lemma}
\begin{proof}
The cases $D=1$ and $d=0$ are immediate.
Both plans minimize
$F(R)=\langle c,R\rangle+\varepsilon\sum R_{jk}\log R_{jk}$
on their respective coupling polytopes.
Lemma~\ref{lem:repair} and the universal distance bound of two give
$R\in\Pi(a,b)$ and $S\in\Pi(\hat a,\hat b)$ with
$\|R-Q\|_1,\|S-P\|_1\le d$.
For probability tables at distance $t\le2$, entropy continuity gives
\[
\begin{aligned}
|H(R_1)-H(R_2)|\le(t/2)\log(D-1)+h(t/2)\le t\log(eD/t),
\end{aligned}
\]
where $h$ is binary entropy. Since $t\log(eD/t)$ is increasing for
$0<t\le2\le D$, either replacement changes the objective by at most
$\Lambda=d\{\bar c+\varepsilon\log(eD/d)\}$.
Optimality of $Q$ yields
\[
F(R)\le F(Q)+\Lambda\le F(S)+\Lambda\le F(P)+2\Lambda.
\]
For $R,P$ with the same marginals, the Gibbs equations give
$F(R)-F(P)=\varepsilon\KL(R\|P)$.
Pinsker's inequality implies
$\|R-P\|_1\le2\sqrt{\Lambda/\varepsilon}$;
the triangle inequality proves the result.
\end{proof}

\begin{theorem}[Finite-sample bound for the transport plan]\label{thm:fss}
Fix retained latent sets, $\varepsilon>0$, and radii $\eta_d,\eta_x$
satisfying \eqref{eq:fss-moment-event} with probability at least
$1-\alpha$. Let $\emin>0$ bound all retained population energies from
below, and assume $\eta_d\le\emin/2$. Define
\[
\begin{gathered}
\delta_c=
\begin{cases}
4(\eta_x+\eta_d)/\emin,&\text{activation cost},\\
0,&\text{fixed decoder cost},
\end{cases}
\\ d_m=\min\{8\eta_d/\emin,2\}.
\end{gathered}
\]
For $0\le c,\hat c\le\bar c$ and $\Ncell=K_AK_B$, set
\[
\Delta=\frac{2\delta_c}{\varepsilon}+d_m+
2\sqrt{\frac{d_m\{\bar c+\varepsilon\log(e\Ncell/d_m)\}}{\varepsilon}},
\]
with square-root term zero when $d_m=0$. Then
\[
\Pr\{\|\hat P-P\|_1\le\min\{2,\Delta\}\}\ge1-\alpha.
\]
One may take $\bar c=2$ for nonnegative activation cosines and
$\bar c=4$ for decoder cosines.
\end{theorem}
\begin{proof}
For $\Ncell=1$, both plans equal $(1)$. Otherwise, on
\eqref{eq:fss-moment-event}, empirical energies are positive and
Lemma~\ref{lem:fss-sens} bounds cost and marginal changes.
Let $Q$ be the optimum for population cost $c$ and empirical marginals
$(\hat a,\hat b)$. Theorem~\ref{thm:stab} gives
$\|\hat P-Q\|_1\le2\delta_c/\varepsilon$.
Lemma~\ref{lem:fss-marg} bounds $\|Q-P\|_1$ by an increasing function
of $d\in[0,2]$, so $d\le d_m$ gives the remaining terms in $\Delta$.
The triangle inequality, the universal distance bound of two, and the
probability of the moment event complete the proof.
\end{proof}

The moment-error radii and energy lower bound determine whether this
guarantee improves on the value two.
\section{Experiments (Section~\ref{sec:exp})}

\subsection{TinyStories Seed Experiment: Reproducibility Details}\label{app:seed-protocol}

\paragraph{Training.}

Training uses 2,048 stories and 4,000 Adam steps (batch 512, learning
rate $3\cdot10^{-4}$, 100-step warmup) to minimize mean squared
reconstruction error. A common root-mean-square of training activations
sets the input scale; codes are restored to original units for energy
moments. Reconstruction is evaluated on 256 tuning stories; the mean
number of active latents is 16.

\paragraph{Animals word list.}
dog, dogs, puppy, puppies, cat, cats, kitten, kittens, rabbit, rabbits, bunny, fox, foxes, bear, bears, lion, lions, tiger, monkey, elephant, bird, birds, duck, horse.

\paragraph{Family word list.}
mother, father, mom, mum, dad, mommy, daddy, parents, sister, brother, grandmother, grandfather, grandma, grandpa, aunt, uncle, cousin.

\paragraph{Computation.}
Training and encoding use FP32 on an NVIDIA L4; moments, transport, and
influence calculations use FP64. Software versions are Python 3.13.16,
PyTorch 2.11.0, NumPy 2.1.3, SciPy 1.16.3, Transformers 4.57.6, and Datasets
3.6.0. Log-domain Sinkhorn allows 20,000 iterations with marginal
$\ell_1$ tolerance $10^{-10}$ and relative tolerance $10^{-7}$.
The adjoint solve fixes one target coordinate and diagonally scales
the reduced system; its condition number and residual are recorded.
The joint covariance of the ten document influence coordinates,
divided by document count, gives pointwise 95\% normal intervals for
mean cost.

The implementation uses $c^{\rm code}=1-\rho$ and
$\varepsilon^{\rm code}\in\{0.05,0.15,0.30\}$. Multiplying both by two
preserves the plans and converts costs, standard errors, and interval
endpoints to \eqref{eq:uncentered-cost}. All 90 transport solves and
20 primary pairwise influence calculations completed. The largest
marginal $\ell_1$ residual was
$9.4\cdot10^{-11}$; scaled adjoint condition numbers were 22.4--99.3.

\paragraph{Full primary matrices.}
Tables~\ref{tab:matrix-animals}--\ref{tab:matrix-family} report
$D_\varepsilon=\langle c,P_\varepsilon\rangle$ for seeds 101--105 at
$\varepsilon=0.30$, including self-costs.

\begin{table}[ht]
\centering\small
\begin{tabular}{lrrrrr}
\toprule
Seed & 101 & 102 & 103 & 104 & 105\\
\midrule
101 & 0.38008 & 1.02088 & 1.05034 & 1.06093 & 1.05751\\
102 & 1.02088 & 0.38572 & 0.97432 & 1.02648 & 1.01152\\
103 & 1.05034 & 0.97432 & 0.38686 & 1.01095 & 1.01947\\
104 & 1.06093 & 1.02648 & 1.01095 & 0.38440 & 1.04600\\
105 & 1.05751 & 1.01152 & 1.01947 & 1.04600 & 0.38763\\
\bottomrule
\end{tabular}
\caption{Animals transport-cost matrix at the primary regularization.}
\label{tab:matrix-animals}
\end{table}

\begin{table}[ht]
\centering\small
\begin{tabular}{lrrrrr}
\toprule
Seed & 101 & 102 & 103 & 104 & 105\\
\midrule
101 & 0.37637 & 1.03810 & 1.07426 & 1.08313 & 1.08056\\
102 & 1.03810 & 0.38514 & 1.00246 & 1.06090 & 1.03596\\
103 & 1.07426 & 1.00246 & 0.38730 & 1.04136 & 1.04432\\
104 & 1.08313 & 1.06090 & 1.04136 & 0.38445 & 1.08132\\
105 & 1.08056 & 1.03596 & 1.04432 & 1.08132 & 0.38644\\
\bottomrule
\end{tabular}
\caption{Family transport-cost matrix at the primary regularization.}
\label{tab:matrix-family}
\end{table}

\subsection{Additional TinyStories-1M Analyses}\label{app:extra-exp}
Five base seeds are used throughout; fifteen more are included in the seed-variation analysis.

\begin{table}[h]
\centering\small
\begin{tabular}{rrrrrr}
\toprule
$\varepsilon$ & $\Delta$ & 95\% interval & diagonal mass & $\sum_j\max_kP_{jk}$ & $\sum_ja_j\exp H(T_{j\cdot})$\\
\midrule
0.03 & $0.02144$ & $[0.01858,0.02429]$ & $1.000$ & $0.674$ & $2.7$\\
0.05 & $0.02161$ & $[0.01873,0.02449]$ & $1.000$ & $0.627$ & $3.3$\\
0.10 & $0.02202$ & $[0.01903,0.02501]$ & $0.998$ & $0.515$ & $5.7$\\
0.20 & $0.02370$ & $[0.02024,0.02715]$ & $0.933$ & $0.344$ & $14.3$\\
0.30 & $0.02636$ & $[0.02273,0.02998]$ & $0.688$ & $0.242$ & $24.0$\\
0.60 & $0.02457$ & $[0.02186,0.02729]$ & $0.208$ & $0.134$ & $39.3$\\
\bottomrule
\end{tabular}
\caption{Contrast over the regularization grid. The self-plan diagonal mass is averaged over seeds and topics. For cross-seed plans, the last two columns give the sum of row maxima and the weighted mean of the exponential row entropy, where $H(T_{j\cdot})=-\sum_kT_{jk}\log T_{jk}$.}
\label{tab:app-eps}
\end{table}

\paragraph{Existing measures.}
Table~\ref{tab:app-baselines} averages ten SAE pairs. Uniform-mass unregularized OT equals Hungarian assignment for these equal-width SAEs. Intervals use 100 document-bootstrap replicates, paired across seeds and independent across topics; positive $z$ denotes lower agreement on \texttt{family}.

\begin{table}[h]
\centering\small
\begin{tabular}{lrrrrr}
\toprule
Score & \texttt{family} & \texttt{animals} & Difference & Bootstrap 95\% & $z$\\
\midrule
Decoder, Hungarian mean cosine & $0.4080$ & $0.4080$ & $+0.0000$ & --- & ---\\
Decoder, nearest-neighbour cosine & $0.4350$ & $0.4350$ & $+0.0000$ & --- & ---\\
Decoder, shared fraction ($\ge0.7$) & $0.0109$ & $0.0109$ & $+0.0000$ & --- & ---\\
Activation $\rho$, Hungarian & $0.5697$ & $0.5701$ & $-0.0005$ & $[-0.0013,0.0003]$ & $1.1$\\
Activation $\rho$, nearest neighbour & $0.6245$ & $0.6262$ & $-0.0017$ & $[-0.0024,-0.0008]$ & $3.6$\\
Activation $\rho$, shared fraction ($\ge0.7$) & $0.2867$ & $0.3086$ & $-0.0219$ & $[-0.0273,-0.0156]$ & $6.9$\\
Pearson correlation, Hungarian & $0.4756$ & $0.4770$ & $-0.0014$ & $[-0.0023,-0.0005]$ & $2.8$\\
Uniform-mass OT, $\varepsilon\to0$ (soft matching) & $0.8606$ & $0.8597$ & $+0.0009$ & $[-0.0006,0.0025]$ & $1.1$\\
Uniform-mass OT, $\varepsilon=0.30$ & $1.2654$ & $1.2640$ & $+0.0014$ & $[-0.0004,0.0030]$ & $1.5$\\
Energy-weighted OT, $\varepsilon\to0$ & $0.7125$ & $0.6911$ & $+0.0213$ & $[0.0185,0.0233]$ & $15.4$\\
Energy-weighted OT, $\varepsilon=0.30$ (ours) & $1.0542$ & $1.0279$ & $+0.0264$ & $[0.0229,0.0291]$ & $15.0$\\
\bottomrule
\end{tabular}
\caption{Seed-agreement scores on the same SAEs and topics.
Higher similarity scores and lower OT costs indicate closer dictionaries.}
\label{tab:app-baselines}
\end{table}

\paragraph{Latents.}
Table~\ref{tab:app-latents} reports the latents in
Figure~\ref{fig:latents}, selected on one half of each topic's documents
and estimated on the other. Energy shares refer to topic-word positions.

\begin{table}[h]
\centering\small
\begin{tabular}{rrrrrrl}
\toprule
Seed & Latent & Contribution & 95\% interval & Family words & Animal words & Top tokens\\
\midrule
105 & 19 & $+0.00153$ & $[0.00139,0.00167]$ & $0.15$ & $0.03$ & \texttt{to} / \texttt{mom} / \texttt{in} / \texttt{my}\\
103 & 43 & $-0.00060$ & $[-0.00089,-0.00031]$ & $0.01$ & $0.03$ & \texttt{.} / \texttt{,} / \texttt{to} / \textbackslash n\\
103 & 33 & $+0.00090$ & $[0.00081,0.00099]$ & $0.17$ & $0.00$ & \texttt{and} / \texttt{mom} / \texttt{he} / \texttt{had}\\
104 & 21 & $+0.00083$ & $[0.00078,0.00088]$ & $0.81$ & $0.00$ & \texttt{mom} / \texttt{my} / \texttt{dad} / \texttt{Mom}\\
101 & 46 & $-0.00092$ & $[-0.00107,-0.00076]$ & $0.01$ & $0.14$ & \texttt{.} / \texttt{it} / \texttt{bird} / \texttt{dog}\\
103 & 23 & $-0.00087$ & $[-0.00100,-0.00074]$ & $0.01$ & $0.03$ & \texttt{a} / \texttt{upon} / \texttt{was} / \texttt{.}\\
101 & 13 & $+0.00077$ & $[0.00070,0.00084]$ & $0.54$ & $0.01$ & \texttt{mom} / \texttt{dad} / \texttt{my} / \texttt{Mom}\\
101 & 1 & $+0.00077$ & $[0.00067,0.00087]$ & $0.01$ & $0.00$ & \texttt{said} / \texttt{day} / \texttt{time} / \texttt{asked}\\
102 & 46 & $+0.00080$ & $[0.00071,0.00089]$ & $0.07$ & $0.00$ & \texttt{to} / \texttt{mom} / \texttt{saw} / \texttt{He}\\
102 & 11 & $+0.00068$ & $[0.00063,0.00074]$ & $0.53$ & $0.00$ & \texttt{mom} / \texttt{my} / \texttt{Mom} / \texttt{dad}\\
\bottomrule
\end{tabular}
\caption{Largest latent contributions to $\Delta$ at $\varepsilon=0.10$.}
\label{tab:app-latents}
\end{table}

\paragraph{Seeds.}
Seeds 106--120 follow the same training recipe and minibatch
stream; tuning FVU across the 20 SAEs is 0.1425--0.1466. We estimate
the standard error of the seed-pair U-statistic by deleting one seed at
a time. All 190 pairs have positive contrasts at
$\varepsilon\in\{0.10,0.30\}$. The 20-seed interval at
$\varepsilon=0.30$ also includes document variance.

\begin{table}[h]
\centering\small
\begin{tabular}{rrrrrr}
\toprule
$\varepsilon$ & Seeds & $\Delta$ & Seed s.e. & Jackknife $t$ interval & Documents\,+\,seeds\\
\midrule
0.10 & 5 & $0.02202$ & $0.00388$ & $[0.01126,0.03278]$ & ---\\
0.10 & 20 & $0.02068$ & $0.00141$ & $[0.01773,0.02364]$ & ---\\
0.30 & 5 & $0.02636$ & $0.00379$ & $[0.01585,0.03687]$ & ---\\
0.30 & 20 & $0.02612$ & $0.00127$ & $[0.02346,0.02877]$ & $[0.02173,0.03050]$\\
\bottomrule
\end{tabular}
\caption{Seed-level uncertainty for the context contrast.}
\label{tab:app-seeds}
\end{table}

\paragraph{Topics.}
Disjoint evaluation pools contain 768 documents each for food, toys, and nature (at least two listed word occurrences) and 1,536 neutral stories (none of the listed words). Family minus neutral is $0.0093$ $[0.0064,0.0122]$; animals minus neutral is $-0.0171$ $[-0.0205,-0.0137]$. Bonferroni intervals exclude zero for 11 of 15 contrasts. Random halves of the neutral pool give $|z|>1.96$ in 2 of 20 splits.

\begin{table}[h]
\centering\small
\begin{tabular}{lrrr}
\toprule
Category & Documents & $M_g$ & 95\% interval\\
\midrule
animals & 768 & $1.02786$ & $[1.02502,1.03070]$\\
nature & 768 & $1.03310$ & $[1.03068,1.03552]$\\
toys & 768 & $1.03584$ & $[1.03346,1.03822]$\\
food & 768 & $1.03732$ & $[1.03485,1.03980]$\\
neutral & 1536 & $1.04494$ & $[1.04310,1.04678]$\\
family & 768 & $1.05422$ & $[1.05197,1.05646]$\\
\bottomrule
\end{tabular}
\caption{Mean transport cost by document category ($\varepsilon=0.30$).}
\label{tab:app-topics}
\end{table}

\paragraph{Coverage.}
Treating each fixed 768-document topic pool as the population, we draw 200 samples of size $m$ with replacement and apply Corollary~\ref{cor:context-cost}. Coverage is assessed against full-pool targets; topics are resampled independently for the contrast.

\begin{table}[h]
\centering\small
\begin{tabular}{lrrrr}
\toprule
Target & $m$ & Coverage & Wilson 95\% & Mean s.e./s.d.\\
\midrule
$M_{\rm family}$ & 192 & $0.945$ & $[0.904,0.969]$ & $1.04$\\
$M_{\rm family}$ & 384 & $0.970$ & $[0.936,0.986]$ & $1.07$\\
$M_{\rm family}$ & 768 & $0.935$ & $[0.892,0.962]$ & $0.94$\\
$M_{\rm animals}$ & 192 & $0.950$ & $[0.910,0.973]$ & $1.01$\\
$M_{\rm animals}$ & 384 & $0.970$ & $[0.936,0.986]$ & $1.05$\\
$M_{\rm animals}$ & 768 & $0.945$ & $[0.904,0.969]$ & $0.98$\\
$\Delta$ & 192 & $0.955$ & $[0.917,0.976]$ & $1.04$\\
$\Delta$ & 384 & $0.970$ & $[0.936,0.986]$ & $1.12$\\
$\Delta$ & 768 & $0.915$ & $[0.868,0.946]$ & $0.94$\\
\bottomrule
\end{tabular}
\caption{Empirical coverage of 95\% intervals ($\varepsilon=0.30$).}
\label{tab:app-coverage}
\end{table}

\subsection{Synthetic Repeated-Sample Calibration}\label{ssec:calibration}

We fix a document distribution with six paired token positions and nonnegative activations for one $3\times4$ SAE pair. Population moments determine the reference plan; we evaluate three fixed fractions and one cost. Support generation uses seed 73 and Monte Carlo sampling uses seed 8103. The regular regime has $\varepsilon=0.20$; a sensitivity regime uses the same support at $\varepsilon=0.04$; a rare-document regime assigns probability 0.005 to a high-impact type.

For each regime and $N\in\{64,256,1024\}$, we draw 200 independent samples (1,800 trials), plus 200 rare-regime samples at $N=4096$. Each sample uses 399 document-bootstrap and 399 Gaussian multiplier draws. The bootstrap recomputes costs and both marginals. We assess
pointwise and Bonferroni Wald intervals, basic bootstrap intervals (centered 0.025/0.975 quantiles), and simultaneous bootstrap and multiplier intervals (0.95 quantile of the maximum absolute deviation
scaled by the original influence standard error). The plan radius is the 0.95 quantile of $\|P^*-\hat P\|_1$. Quantiles use linear
interpolation; 399 draws entail Monte Carlo quantile error.

\begin{table}[t]
\centering\small
\begin{tabular}{rrrr}
\toprule
$N$ & Mean plan error & Mean radius & Coverage\\
\midrule
64 & 0.07922 & 0.12017 & 0.965\\
256 & 0.04081 & 0.06048 & 0.955\\
1024 & 0.02024 & 0.03019 & 0.970\\
\bottomrule
\end{tabular}
\caption{Regular finite-support regime: full-plan L1 error and
95\% document-bootstrap radius. Coverage is the fraction of 200 independent document samples for which the confidence region contains the population plan.}
\label{tab:bootstrap-radius}
\end{table}

Table~\ref{tab:bootstrap-radius} shows approximately $N^{-1/2}$
scaling. Simultaneous coverage for the three fractions and one cost is
0.945, 0.945, 0.955 at $N=64,256,1024$ with bootstrap maxima and
0.955, 0.950, 0.965 with multiplier maxima. Basic-bootstrap and Wald
cost coverage both equal 0.940 at $N=256$ and 0.950 at $N=1024$.

The energy-error condition for the finite-sample bound of Appendix~\ref{app:fss} holds only in the regular and small-$\varepsilon$ regimes at $N=1024$, where its plan radius is two. Elsewhere two is the universal $\ell_1$ diameter of probability plans; the bound is uninformative.

In the rare-document regime, full-plan coverage is
0.275, 0.720, 0.850, 0.935 at $N=64,256,1024,4096$; simultaneous
fraction-and-cost coverage is 0.275, 0.720, 0.925, 0.935. At $N=64$,
145 of 200 samples miss the rare type, and all 145 simultaneous bands
fail. With $\varepsilon=0.04$, simultaneous coverage is
0.905, 0.920, 0.960 at $N=64,256,1024$.

\subsection{GPT-2 Replication}\label{app:gpt2}
For \texttt{openai-community/gpt2}, we use the block 6 output (zero-based, $d=768$) and the same corpus, topics, prefixes, data seed, and hash split as TinyStories-1M. Five TopK SAEs per width (3072 and 6144) use $k=32$, 24,576 training stories.
\begin{table}[h]
\centering\small
\begin{tabular}{rrrrr}
\toprule
Width & $\varepsilon$ & $\Delta$ & 95\% interval & self-plan diagonal\\
\midrule
3072 & $0.1$ & $-0.0003$ & $[-0.0021,0.0016]$ & $0.81$\\
3072 & $0.2$ & $+0.0012$ & $[-0.0024,0.0048]$ & $0.64$\\
3072 & $0.3$ & $+0.0056$ & $[0.0007,0.0105]$ & $0.34$\\
3072 & $0.6$ & $+0.0084$ & $[0.0046,0.0123]$ & $0.13$\\
6144 & $0.1$ & $+0.0050$ & $[0.0033,0.0067]$ & $0.77$\\
6144 & $0.2$ & $+0.0080$ & $[0.0042,0.0118]$ & $0.58$\\
6144 & $0.3$ & $+0.0069$ & $[0.0020,0.0117]$ & $0.30$\\
6144 & $0.6$ & $+0.0072$ & $[0.0035,0.0110]$ & $0.12$\\
\bottomrule
\end{tabular}
\caption{GPT-2 context contrast by width and regularization.}
\label{tab:app-gpt2}
\end{table}

At width 6144, all ten seed-pair contrasts are positive at $\varepsilon=0.10,0.20$ (Table~\ref{tab:app-gpt2}). At $\varepsilon=0.10$, the cost-matrix components are $+0.0027$ $[+0.0016,+0.0037]$ and $+0.0037$ $[+0.0027,+0.0048]$ at widths 3072 and 6144; the weight components are $-0.0029$ $[-0.0044,-0.0015]$ and $+0.0014$ $[+0.0000,+0.0028]$. With ten seeds at width 3072, the contrast is 0.0011 with combined interval $[-0.0014,0.0035]$. Seed-variation analysis covers only
width 3072. Excluding topic-word positions leaves its interval spanning zero and reduces the width-6144 contrast to 0.0032 $[0.0015,0.0049]$.

\subsection{Pythia and TinyStories-33M: Comparison Across Models}
\label{app:larger-models}

\paragraph{Common protocol.}
Pythia-160M (block 6 output) and TinyStories-33M (block 1 output), both $d=768$, follow the GPT-2 protocol (Appendix~\ref{app:gpt2}). Each has five SAEs (seeds 101--105) at widths 3072 and 6144, sharing training documents and minibatches within a setting. Evaluation uses 768 documents per topic and 256 neutral tuning documents. The main regularization is $\varepsilon=0.10$, with additional values 0.20, 0.30, and 0.60. Comparisons use independently initialized SAEs of equal width; seeds 106--110 extend the width-3072 seed analysis. Table~\ref{tab:app-model-comparison} summarizes the seven settings.
\begin{table}[ht]
\centering\small
\begin{tabular}{lrrrrrr}
\toprule
Model & Width & Tuning FVU & \shortstack{Self\\diagonal} & $\Delta$ & \shortstack{Positive\\pairs} & \shortstack{Fraction\\tests}\\
\midrule
TinyStories-1M & 64 & $0.143$--$0.147$ & $1.00/0.69$ & $+0.0220$ & $10/10$ & $91/120$\\
TinyStories-33M & 3072 & $0.180$ & $0.76/0.36$ & $-0.0007$ & $4/10$ & $55/120$\\
TinyStories-33M & 6144 & $0.151$--$0.152$ & $0.73/0.37$ & $+0.0021$ & $8/10$ & $73/120$\\
GPT-2 small & 3072 & $0.088$--$0.089$ & $0.81/0.34$ & $-0.0003$ & $5/10$ & $70/120$\\
GPT-2 small & 6144 & $0.073$ & $0.77/0.30$ & $+0.0050$ & $10/10$ & $62/120$\\
Pythia-160M & 3072 & $0.018$ & $0.29/0.15$ & $-0.0026$ & $3/10$ & $69/120$\\
Pythia-160M & 6144 & $0.015$ & $0.28/0.15$ & $-0.0011$ & $5/10$ & $62/120$\\
\bottomrule
\end{tabular}
\caption{Five-seed comparisons on identical evaluation topics. Self-plan diagonal masses are shown at $\varepsilon=0.10/0.30$; all other transport quantities use $\varepsilon=0.10$. Positive pairs counts the seed pairs with $\Delta>0$. Fraction tests counts transported-fraction differences between topics significant after Bonferroni correction within the 120
comparisons for that setting. FVU and contrast values are rounded.}
\label{tab:app-model-comparison}
\end{table}

\paragraph{Context contrasts and seed variation.}
At $\varepsilon=0.10$, contrasts for the six larger
settings range from $-2.3\%$ to $+2.1\%$ of the animals mean cost. Cost-matrix components are positive in GPT-2 and Pythia (approximately $0.9$--$1.5\%$ of that mean), and near zero in TinyStories-33M; weight components have either sign. Width 6144 has a larger contrast than width 3072 in each model at $\varepsilon=0.10,0.20$. Ten-seed intervals at width 3072 include zero for all three models (Table~\ref{tab:app-large-seeds}).

\begin{table}[ht]
\centering\small
\begin{tabular}{lrr}
\toprule
Model & $\Delta$, ten seeds & 95\% interval\\
\midrule
TinyStories-33M & $+0.0004$ & $[-0.0023,+0.0032]$\\
GPT-2 small & $+0.0011$ & $[-0.0014,+0.0035]$\\
Pythia-160M & $-0.0016$ & $[-0.0049,+0.0016]$\\
\bottomrule
\end{tabular}
\caption{Ten-seed context contrasts at width 3072 and
$\varepsilon=0.10$. The analysis includes variation between training
seeds, assessed by deleting one seed at a time; the GPT-2 interval also
includes document variation as described in Appendix~\ref{app:gpt2}.}
\label{tab:app-large-seeds}
\end{table}

\paragraph{Regularization and topic-word positions.}
At $\varepsilon=0.60$, all six larger settings have positive document intervals, with contrasts of $0.0036$--$0.0084$. Self-plan diagonal mass is at most $0.37$ at $\varepsilon=0.30$; at $\varepsilon=0.10$ it is $0.28$--$0.29$ in Pythia and $0.73$--$0.81$ in GPT-2 and TinyStories-33M. Table~\ref{tab:app-word-exclusion} recomputes moments with topic-word positions excluded. The activations still depend on their original context; only the positions used in the moments change.

\begin{table}[ht]
\centering\small
\begin{tabular}{lrrr}
\toprule
Model & Width & Original $\Delta$ & Word positions excluded\\
\midrule
TinyStories-1M & 64 & $+0.0220$ & $+0.0179$\\
TinyStories-33M & 3072 & $-0.0007$ & $-0.0011$\\
TinyStories-33M & 6144 & $+0.0021$ & $+0.0022$\\
GPT-2 small & 3072 & $-0.0003$ & $-0.0005$\\
GPT-2 small & 6144 & $+0.0050$ & $+0.0032$\\
Pythia-160M & 3072 & $-0.0026$ & $-0.0027$\\
Pythia-160M & 6144 & $-0.0011$ & $-0.0013$\\
\bottomrule
\end{tabular}
\caption{Effect of excluding topic-word positions, using the five-seed
runs at $\varepsilon=0.10$. The TinyStories-1M value retains approximately
81\% of the contrast; the 78\% figure in the main text uses
$\varepsilon=0.30$.}
\label{tab:app-word-exclusion}
\end{table}

\subsection{Decomposing Disagreement by Activation Energy}
\label{app:energy-groups}

\paragraph{Group definition and evaluation.}
Rank each SAE's latents by uncentered activation energy on 256 neutral tuning documents. The smallest prefix reaching 50\% of energy forms the high-energy group; the next latents up to 90\% form the middle group; the rest form the low-energy group. Whole latents yield approximate shares. Groups are fixed before evaluation.

\paragraph{Cost shares.}
In all 35 SAEs, the high-energy group's cost share is below its energy share and the low-energy group's is above it (Table~\ref{tab:app-energy-groups}). Across the six larger settings, five-SAE averages assign 16--29\% of evaluation cost to groups selected for about 50\% of tuning energy and 28--40\% to groups selected for about 10\%.

\begin{table}[ht]
\centering\small
\begin{tabular}{lrrr}
\toprule
Settings & SAEs & \shortstack{High-energy group\\cost share / energy share} & \shortstack{Low-energy group\\cost share / energy share}\\
\midrule
TinyStories-1M, width 64 & 5 & $0.84$--$0.89$ & $1.3$--$1.4$\\
Three larger models, two widths each & 30 & $0.11$--$0.68$ & $2.4$--$4.5$\\
\bottomrule
\end{tabular}
\caption{Descriptive cost-share to energy-share ratios across SAEs.
Groups are selected on neutral tuning documents; cost is evaluated at
$\varepsilon=0.10$. Ratios below one indicate less cost per unit energy
than the dictionary average.}
\label{tab:app-energy-groups}
\end{table}

High-energy groups contain 14--18 latents in TinyStories-1M, 2--3 in TinyStories-33M, 6--11 in GPT-2, and 12--16 in Pythia across their tested widths. In larger models, they fire in 192--256 of 256 tuning documents, often on punctuation and function words. Their contrast contribution is negative in five of seven settings, offsetting
other latents in GPT-2 at width 3072 and accounting for most of Pythia's negative contrast.

\end{document}